\documentclass[acmlarge]{acmart}
\AtBeginDocument{%
  }

\setcopyright{acmlicensed}
\copyrightyear{2026}
\acmYear{2026}
\acmDOI{XXXXXXX.XXXXXXX}

\acmJournal{POMACS}
\acmVolume{37}
\acmNumber{4}
\acmArticle{111}
\acmMonth{8}

\usepackage{url}
\usepackage{xcolor}
\usepackage{graphicx}

\usepackage{longtable}

\usepackage{hyperref}
\usepackage{multirow}
\usepackage{algorithmic}
\usepackage{textcomp}
\usepackage{multirow,multicol, makecell, booktabs}

\usepackage{graphicx}
\usepackage{float}
\usepackage{booktabs, makecell, tabularx}
\usepackage{rotating}
\usepackage[export]{adjustbox}
\usepackage{booktabs}
\usepackage{subcaption}
\usepackage{graphicx}
\usepackage{subcaption} 
\usepackage{pifont}

\definecolor{newcolor}{rgb}{.8,.349,.1}

\begin{document}

\title{Recent Advances in Medical Imaging Segmentation: A Survey}

\author{Fares Bougourzi}
\email{faresbougourzi@gmail.com}
\orcid{0000-0001-5077-4862}
\affiliation{%
\institution{IEMN, UPHF, University of Lille}
\state{Valenciennes}
\country{France}
}

\author{Abdenour Hadid}
\authornote{This denotes the corresponding author.}
\email{abdenour.hadid@ieee.org}
\orcid{0000-0001-9092-735X}
\affiliation{%
  \institution{Faculty of Data Science and Computing, University Malaysia Kelantan}
  \state{Kelantan} 
  \country{Malaysia}
}

\renewcommand{\shortauthors}{Bougourzi et al.}

\begin{abstract}
Medical imaging is a cornerstone of modern healthcare, driving advancements in diagnosis, treatment planning, and patient care. Among its various tasks, segmentation remains one of the most challenging problem due to factors such as data accessibility, annotation complexity, structural variability, variation in medical imaging modalities, and privacy constraints. Despite recent progress, achieving robust generalization and domain adaptation remains a significant hurdle, particularly given the resource-intensive nature of some proposed models and their reliance on domain expertise.
This survey explores cutting-edge advancements in medical image segmentation, focusing on methodologies such as Generative AI, Few-Shot Learning, Foundation Models, and Universal Models. These approaches offer promising solutions to longstanding challenges. We provide a comprehensive overview of the theoretical foundations, state-of-the-art techniques, and recent applications of these methods. Finally, we discuss inherent limitations, unresolved issues, and future research directions aimed at enhancing the practicality and accessibility of segmentation models in medical imaging. We are maintaining a \href{https://github.com/faresbougourzi/Awesome-DL-for-Medical-Imaging-Segmentation}{GitHub Repository} to continue tracking and updating innovations in this field. 
\end{abstract}



\received{8 July 2025}
\received[revised]{24 June 2026}
\received[accepted]{}

\maketitle

\section{Introduction}
\label{sec:introduction}
Medical Image Segmentation (MIS) is a critical task in medical imaging analysis, playing a key role in various clinical applications such as computer-aided diagnosis, treatment planning, and disease progression monitoring \cite{weese2016four, hansen2022anomaly, ma2024segment}. Performing medical image segmentation typically requires experienced radiologists and substantial time and effort for annotation \cite{weese2016four, hansen2022anomaly, ma2024segment, zhang2024segment}. Over the past decade, Machine Learning (ML), particularly Deep Learning (DL), has demonstrated significant potential in this field, achieving expert-level performance when sufficient and well-labeled datasets are available. Despite these advancements, developing an efficient machine learning model for real-world applications remains a challenging task, as MIS inherits difficulties from both medical and computer vision fields \cite{weese2016four, hansen2022anomaly, ma2024segment, bougourzi2023pdatt}.

From a medical perspective, data acquisition is highly constrained due to privacy concerns and the need for expert annotations, which are time-consuming and labor-intensive. Furthermore, the diversity of medical imaging modalities, each with distinct appearances and properties, adds another layer of complexity (Figure \ref{fig:mod}). In contrast to natural image segmentation, medical image segmentation faces additional challenges, including target structures with varying shapes, weak boundaries, and low contrast, necessitating more robust approaches \cite{weese2016four, hansen2022anomaly, ma2024segment}.

From a machine learning standpoint, deep learning models struggle with generalization across datasets due to differences in scanning device characteristics and recording settings \cite{hansen2022anomaly, ma2024segment}. Additionally, segmenting new anatomical structures typically requires training a new model with sufficient labeled data, often demanding ML expertise, thereby limiting the accessibility of deep learning for medical professionals \cite{hansen2022anomaly, ma2024segment, zhang2024segment}.

To address these challenges, various advanced deep learning methods have been proposed, yielding promising results. This survey investigates recent developments in these methods, analyzing their strengths and limitations. Unlike existing surveys on medical image segmentation, which primarily focus on architectural designs and loss functions \cite{siddique2021u, azad2024medical} or provide a broad overview of Transformer-based advancements across multiple medical imaging tasks, including segmentation \cite{azad2024advances, shamshad2023transformers, li2023transforming}, our work takes a more targeted approach. Specifically, we go beyond these aspects by exploring cutting-edge methodologies designed to tackle the unique challenges of medical image segmentation. In particular, we emphasize recent innovations in generative models, few-shot learning, foundation models, and universal models, providing a comprehensive perspective on their potential to advance the field.

Generative models are designed to generate realistic samples to cope with the lack of sufficient data, privacy constraints, and differences in medical imaging modalities. These models are particularly suitable for cross-modal exploitation in the medical imaging field, addressing challenges such as data scarcity and modality variations \cite{huo2018synseg, sandfort2019data, wu2024medsegdiff2}. While generative models focus on creating synthetic data to augment training, Few-Shot Segmentation (FSS) methods tackle the challenge of learning new tasks with minimal labeled data. FSS reduces the dependence on extensive annotations and retraining efforts, making it a practical solution for scenarios where labeled data is scarce or costly to obtain \cite{hansen2022anomaly, zhang2024prototype}.

Similarly, foundation models have demonstrated remarkable potential in reducing the reliance on tedious manual annotations by leveraging prompt-based segmentation techniques, such as point-based, bounding-box-based, and text-based prompts. Beyond obtaining accurate segmentation masks, these models have proven useful in segmentation label preparation, generalization across domains, and cross-modality segmentation \cite{ma2024segment, zhang2024segment}. Recently, universal models have gained increasing attention in the research community due to their ability to handle domain shifts and adapt to new segmentation tasks with minimal examples, further pushing the boundaries of model generalization and applicability \cite{liu2023clip}.

This survey aims to provide a comprehensive analysis of these emerging methodologies, positioning them within the broader context of medical image segmentation. Specifically, we:


\begin{figure*}
\centering
\includegraphics[width=1.00\textwidth]{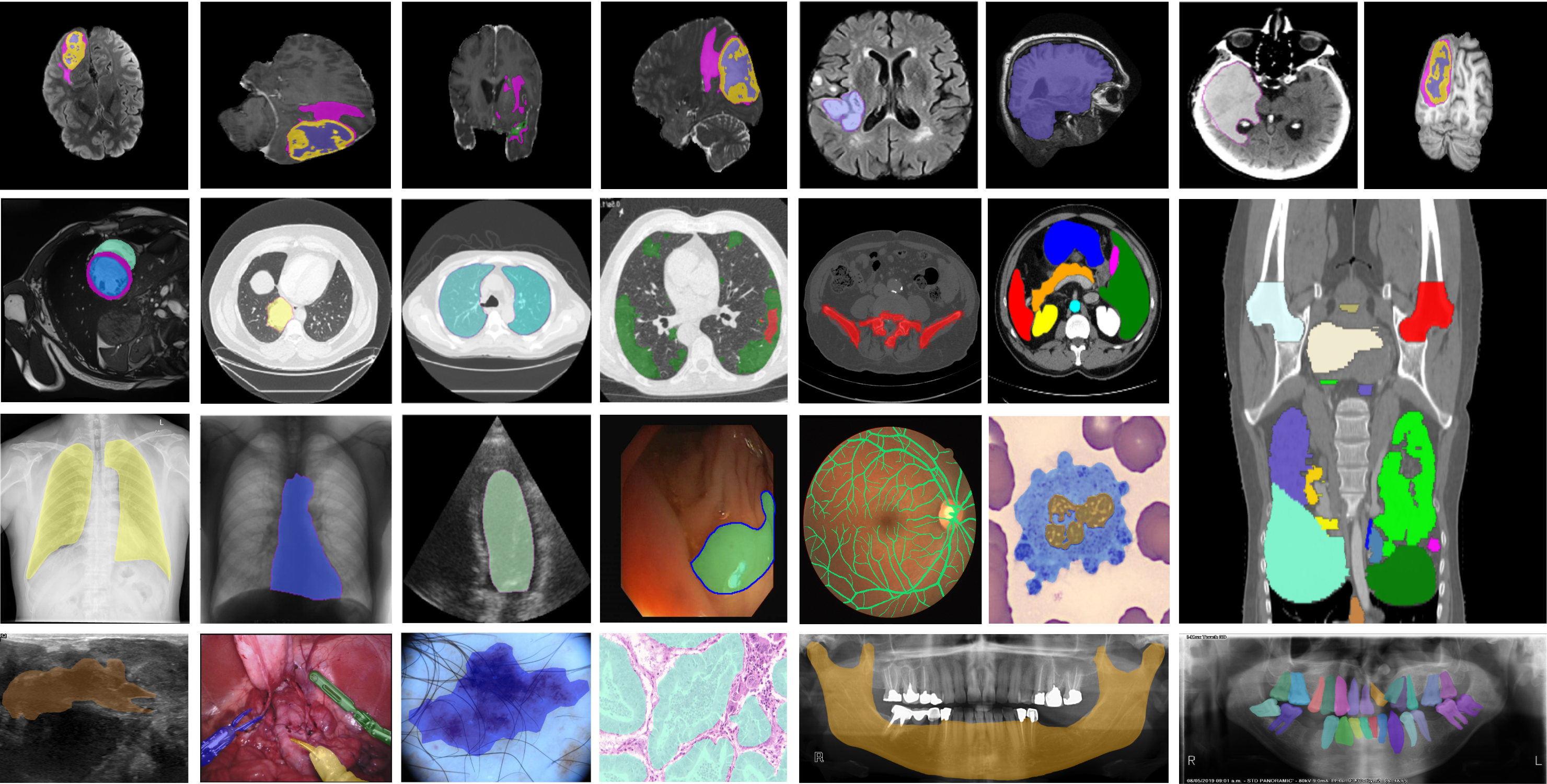}
\caption{Examples of segmentation modalities and tasks in medical imaging. The figure shows the diversity of medical imaging modalities, each with distinct appearances and properties including varying shapes, weak boundaries, and low contrast.}
\label{fig:mod}
\end{figure*}

\begin{itemize}
    \item Provide an in-depth discussion of the core challenges in MIS from both medical and machine learning perspectives.
    \item Present a detailed review of generative models, few-shot learning, foundation models, and universal models, analysing their contributions and potential impact.
    \item Compare the strengths and limitations of these approaches, highlighting their effectiveness in real-world clinical applications.
    \item Identify open research questions and future directions, shedding light on potential pathways for improving MIS models and their integration into medical workflows.
\end{itemize}

By synthesizing the latest advancements, this survey can serve as a valuable resource for researchers and practitioners aiming to develop robust and efficient medical image segmentation models, bridging the gap between theoretical research and practical deployment in clinical settings. 

The remainder of this paper is organized as follows: Section \ref{sec:cha} summarizes the challenges in medical image segmentation and highlights recent advancements. In Section \ref{Sec:GA}, we discuss the role of generative AI in MIS. Section \ref{Sec:FFS} is devoted to few-shot segmentation (FSS). Sections \ref{Sec:FM} and \ref{Sec:UM} highlight advancements in foundation and universal models, respectively. In sections \ref{Sec:dis}, we discuss the current progress in these methods and outline potential directions for future work. Finally, we conclude the survey in Section \ref{Sec:Conc}.

\section{Challenges in Medical Imaging Segmentation}
\label{sec:cha}
\begin{figure}[htp]
\centering
\includegraphics[width=0.7\textwidth]{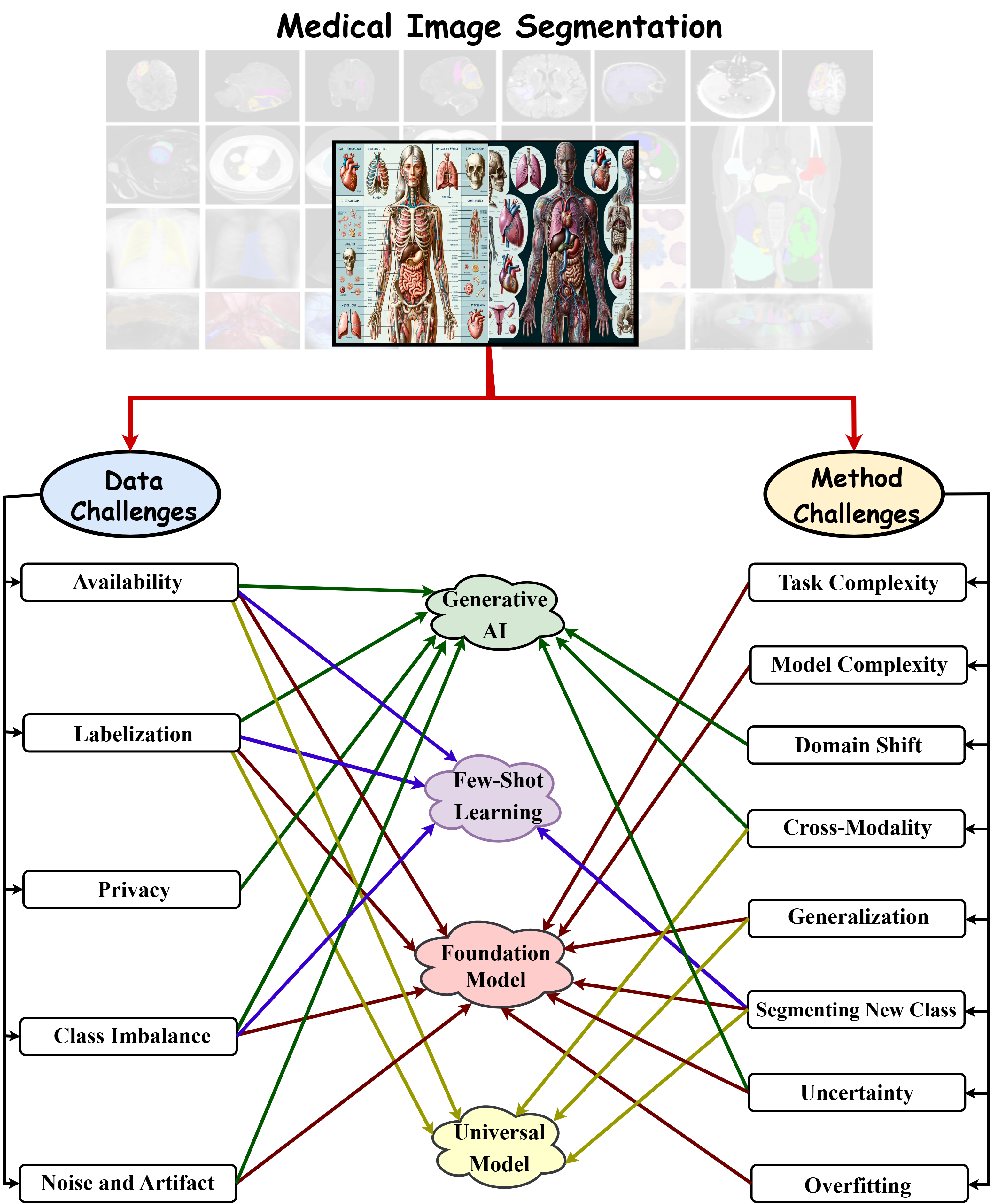}
\caption{General overview of MIS challenges. This survey focuses on the latest developments in generative AI models, few-shot segmentation techniques, foundation models, and universal models. These approaches have demonstrated their effectiveness in addressing key challenges in medical image segmentation, including both data-related and method-specific challenges.}
\label{fig:cha}
\end{figure}

Medical imaging segmentation (MIS) is one of the most challenging tasks in computer vision, posing significant hurdles in both data preparation and the development of efficient methods. As illustrated in Figure \ref{fig:cha}, these challenges can be broadly categorized into two main groups: (i) data-related challenges and (ii) method-based challenges \cite{weese2016four, bougourzi2023pdatt}.

Preparing an adequate and well-constructed dataset for training machine learning models involves a range of difficulties, including data availability, privacy constraints, class imbalance, and noise or artifacts \cite{weese2016four, bougourzi2023pdatt, azad2024medical, huo2018synseg, sandfort2019data}. Obtaining sufficient medical imaging data is often a tedious and time-consuming process that can span months or even years. This difficulty is exacerbated by strict privacy regulations that limit data sharing and accessibility \cite{huo2018synseg, sandfort2019data, wu2024medsegdiff2}. Even when data is available, generating high-quality annotations for segmentation requires domain expertise and significant effort, as the process involves pixel-level labeling. This manual annotation process can be subject to expert subjectivity, leading to variability across different annotators. To mitigate this issue and achieve reliable annotations, multiple experts are often required to verify and refine the labels \cite{weese2016four, bougourzi2023pdatt, azad2024medical, wu2024medsegdiff2}.

Another critical challenge is class imbalance, where different classes (e.g., tissues, abnormalities, infections) have varying sizes and frequencies of occurrence across cases \cite{weese2016four, bougourzi2023pdatt}. Imbalanced datasets can bias the learning process, leading to suboptimal model performance on underrepresented classes~\cite{weese2016four, bougourzi2023pdatt}. Additionally, medical images frequently suffer from noise and artifacts, such as low contrast, weak boundaries, or acquisition-related distortions, which necessitate multi-stage preprocessing and filtering to ensure data quality.

Developing machine learning approaches that are both efficient and applicable in real-world medical scenarios introduces a range of methodological challenges, including task complexity, model generalization, domain shift, uncertainty, and scalability.
Medical imaging segmentation is inherently more complex than natural image segmentation due to the variety of anatomical structures and their diverse appearance mechanisms. Objects in medical images may include nominal body tissues, microscopic abnormalities, infections, and lesions, all exhibiting varying shapes, weak boundaries, and low contrast. Additionally, medical imaging encompasses a broad range of modalities (e.g., CT, MRI, ultrasound), each with specific advantages, drawbacks, and acquisition conditions. The variability in imaging devices, recording settings, and patient populations further exacerbates the domain shift problem, where the performance of a model trained on one dataset may degrade when applied to another.

Generalization is another critical challenge, as models must perform well across diverse datasets and imaging conditions despite differences in data distribution between source and target domains \cite{hansen2022anomaly, ma2024segment, zhang2024segment}. In some cases, there may be abundant labeled data from a source modality but little or no labeled data for the target modality  \cite{huo2018synseg, sandfort2019data}. Developing models capable of exploiting source data while effectively adapting to new target domains is needed. 

The uncertainty of model predictions is also a notable concern, particularly when segmentation labels are influenced by factors such as expert subjectivity, low-contrast boundaries, or label noise \cite{rakic2024tyche}. These issues can lead to considerable variation in annotations, ultimately affecting the confidence and reliability of the model's predictions. Handling this uncertainty is crucial for producing robust and trustworthy segmentation results.

Moreover, deep learning-based methods often involve large-scale models with millions of trainable parameters, which require substantial training data to avoid overfitting. In scenarios with insufficient data, models tend to overfit the training set, leading to a drop in performance on unseen samples \cite{ma2024segment, zhang2024segment}. This challenge is further amplified when attempting to segment new classes that were not part of the original training set. Adding new classes typically requires retraining the model with additional labeled data, a process that demands significant computational resources and machine learning expertise, resources that may not be readily available to medical professionals \cite{ma2024segment, zhang2024segment}.

Finally, while recent advances in deep learning have led to highly accurate models, these methods are often computationally expensive and complex to implement. This limits their usability for medical practitioners who require lightweight, user-friendly models that balance performance and complexity.

To tackle these numerous and varied challenges, recent advancements in deep learning have introduced innovative methods with promising potential. In this survey, we focus on the latest developments in generative AI models, few-shot segmentation techniques, foundation models, and universal models. These approaches have been selected for their ability to address critical MIS challenges, such as limited data availability, domain shift, uncertainty, and the need for scalable, efficient solutions. As summarized in Figure \ref{fig:cha}, these methods represent the forefront of research in medical image segmentation and offer new avenues for overcoming longstanding obstacles in the field.

\section{Genrative AI}
\label{Sec:GA}
Deep Generative Models (DGMs) have shown great capability in generating realistic samples in computer vision. These models including GANs \cite{Radford2015UnsupervisedRL}, Diffusion Models \cite{pmlr-v37-sohl-dickstein15} Variational Autoencoders \cite{Kingma2013AutoEncodingVB} and Normalizing Flows \cite{pmlr-v37-rezende15}, have a wide range of applications. In medical imaging segmentation, the most used approaches are GANs and diffusion models.

\subsection{Theoretical Background}
In this section, a brief introduction to the concepts for GANs and diffusion models are provided to facilitate the understanding of current progress in medical imaging segmentation field. For more details, the reader can refer to \cite{bond2021deep, croitoru2023diffusion, prince2023understanding}.

\begin{figure}[htp]
\centering
\includegraphics[width = 5.5in, height=5in]{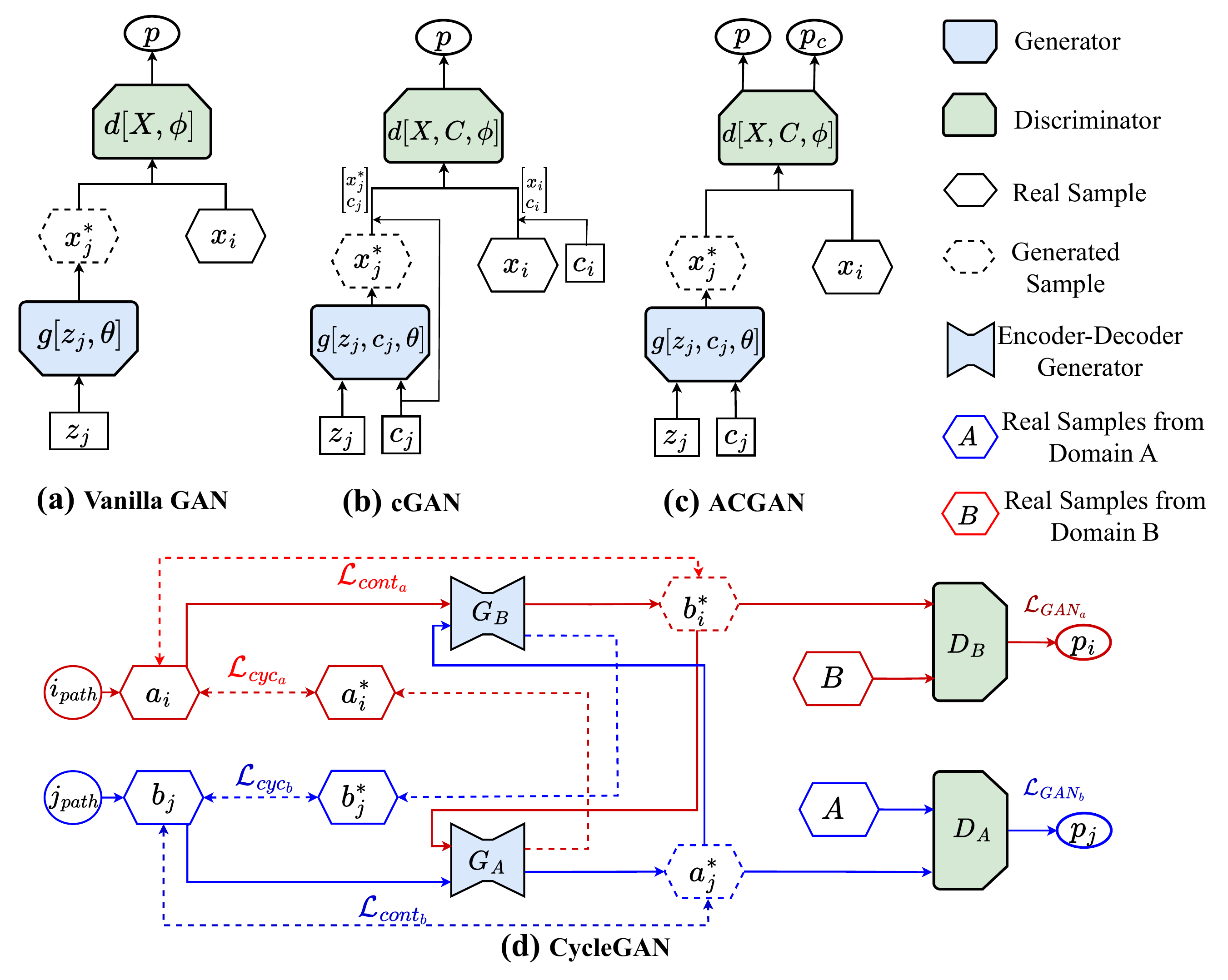}
\caption{Illustration of generative adversarial models, including: (a) Vanilla GAN \cite{goodfellow2014generative}, (b) cGAN \cite{mirza2014conditional}, (c) ACGAN \cite{odena2017conditional} and (d) CycleGAN \cite{karras2019style}.}
\label{fig:gan}
\end{figure}

\textbf{GANs}: Generative Adversarial Networks are unsupervised methods that allow to generate new sample images that are not distinguishable from the real images used during the training (Figure \ref{fig:gan}.a) \cite{goodfellow2014generative, prince2023understanding}. In summary, GANs consists of two networks called \textit{Generator (G)} and \textit{Discriminator (D)}. The Generator with parameters $\theta$ generates new sample $x^*_j$ using latent variable $z_j$ from simple base distribution; $x^*_j = g[z_j, \theta]$.  On the other hand, the aim of the discriminator with parameters $\phi$ is to distinguish between the real data samples $x_i$ and the generated samples; $p = d[X, \phi]$, where $X$ is constructed by real and generated image samples and $p$ is the prediction if $X$ is real or fake (generated). The training of GAN model is also known as \textit{minimax game}. In more details, the objective function of D is to find the optimal parameters $\phi$ that minimize the following function:

\begin{equation}
\scalebox{0.8}{$
\hat{\phi} = \underset{\phi}{\arg\min} \left[ \underset{j}{\sum} -\log ( 1 - \text{sig} (d[x^*, \phi])  ) \right. 
\left. - \underset{i}{\sum} \log \left( \text{sig} \left( d[x, \phi] \right) \right) \right]
$}
\end{equation}

where $sig$ is the logistic sigmoid function. On the other hand, the objective function of G is to find the optimal parameters $\theta$ that maximize the following function:

\begin{equation}
\scalebox{0.7}{$
\hat{\theta} = \underset{\theta}{\arg\max} \left[ \underset{\phi}{\min} \left[ \underset{j}{\sum} -\log \left( 1 - \text{sig} \left( d[g[z_j, \theta], \phi] \right) \right) \right. \right. 
\left. \left. - \underset{i}{\sum} \log \left( \text{sig} \left( d[x, \phi] \right) \right) \right] \right]
$}
\end{equation}


This means that G is optimized to find the parameters that are able to generate samples that the discriminator will not be able to distinguish from the real ones. Thus, the discriminator loss function is binary cross-entropy as follow:

\begin{equation}
\scalebox{0.9}{$
{\mathcal{L}}(\phi) =   \underset{j} {\sum{}} -log ( 1- sig (d[g[z_j, \theta], \phi] )) -
\underset{i} {\sum{}} log ( sig (d[x, \phi] ) ) $}  \end{equation}

On the other hand, the loss function of the generator is the multiplication of the first term of $\mathcal{L}(\phi)$ by $-1$ to encourage the network to generate sample similar to the real ones and removing the second part, which is independent from the parameters $\theta$. Thus, G loss function is defined by:
\begin{equation}
{\mathcal{L}}(\theta) =   \underset{j} {\sum{}}log ( 1- sig (d[g[z_j, \theta], \phi] ))      
\end{equation}

Patch-based discriminators have been widely adopted \cite{li2016precomputed, zhu2017unpaired}. Instead of assigning a single binary classification to the entire image as real or fake, these discriminators assign binary classifications (real/fake) to small patches. This approach sharpens the model's focus on local textures and finer details, resulting in the generation of more realistic and highly detailed images.

\textbf{Conditional GAN:} While Vanilla GANs can generate realistic images, they lack control over specific attributes of those images. Conditional GANs address this limitation by conditioning both the generator and discriminator on desired attributes \(c_j\) \cite{mirza2014conditional}. These attributes can be incorporated in various ways, such as embedding features (in standard cGANs \cite{mirza2014conditional}, Figures \ref{fig:gan}.b) or through an auxiliary classification task (in ACGANs \cite{odena2017conditional}, Figure \ref{fig:gan}.c). In cGAN, the embedded features representing the attributes are fed to both the generator and the discriminator, allowing the model to generate images that conform to the specified attributes. In contrast in ACGAN,  the attributes are fed to the generator as prior knowledge and introduces and used as additional classification loss in the discriminator, alongside the standard adversarial loss. This setup enables both cGANs and ACGANs to produce images that not only appear realistic but also conform to specific, predefined attributes.

\textbf{CycleGAN}: In addition to generating images, GANs have shown great potential for image-to-image translation and style transfer \cite{karras2019style}, as shown in Figure \ref{fig:gan}.d. CycleGAN consists of two generators ($G_A$ and $G_B$) and two discriminators ($D_A$ and $D_B$) that translate an image $a$ from domain $A$ to an image $b^*$ in domain $B$, and also perform the reverse process. The process involves two paths: In the first path $i$, the generator $G_B$ translates a real image $a_i$ from domain $A$ into a corresponding image $b^*_i$ in domain $B$. Then, the generator $G_A$ translates the image $b^*_i$ back into the original domain $A$, producing a reconstructed image $a^*_i$. In the second path $j$, which is the reverse process, the generator $G_A$ translates a real image $b_j$ from domain $B$ into a corresponding image $a^*_j \in A$. The generator $G_B$ then translates $a^*_j$ back into domain $B$, producing a reconstructed image $b^*_j$. The discriminators $D_B$ and $D_A$ are trained to distinguish between generated and real images from domains $B$ and $A$, respectively. CycleGAN is trained using a weighted sum of three losses: adversarial loss ($\mathcal{L}_{GAN_a}$ and $\mathcal{L}_{GAN_b}$), content loss ($\mathcal{L}_{cont_a}$ and $\mathcal{L}_{cont_b}$), and cycle consistency loss ($\mathcal{L}_{cyc_a}$ and $\mathcal{L}_{cyc_b}$). The adversarial loss is the standard GAN loss that encourages each generator to produce images indistinguishable from real images in the target domain. The content loss is typically an $L_1$ norm loss between the input and output of the generator, which aims to preserve the general content of the input in the translated image. The cycle consistency loss ensures that the input image can be reconstructed after a round-trip translation (e.g., $a_i \rightarrow b^*_i \rightarrow a^*_i$ and $b_j \rightarrow a^*_j \rightarrow b^*_j$) \cite{zhu2017unpaired}.

\begin{figure}[htp]
\centering
\includegraphics[width = 5.5in, height=5in]{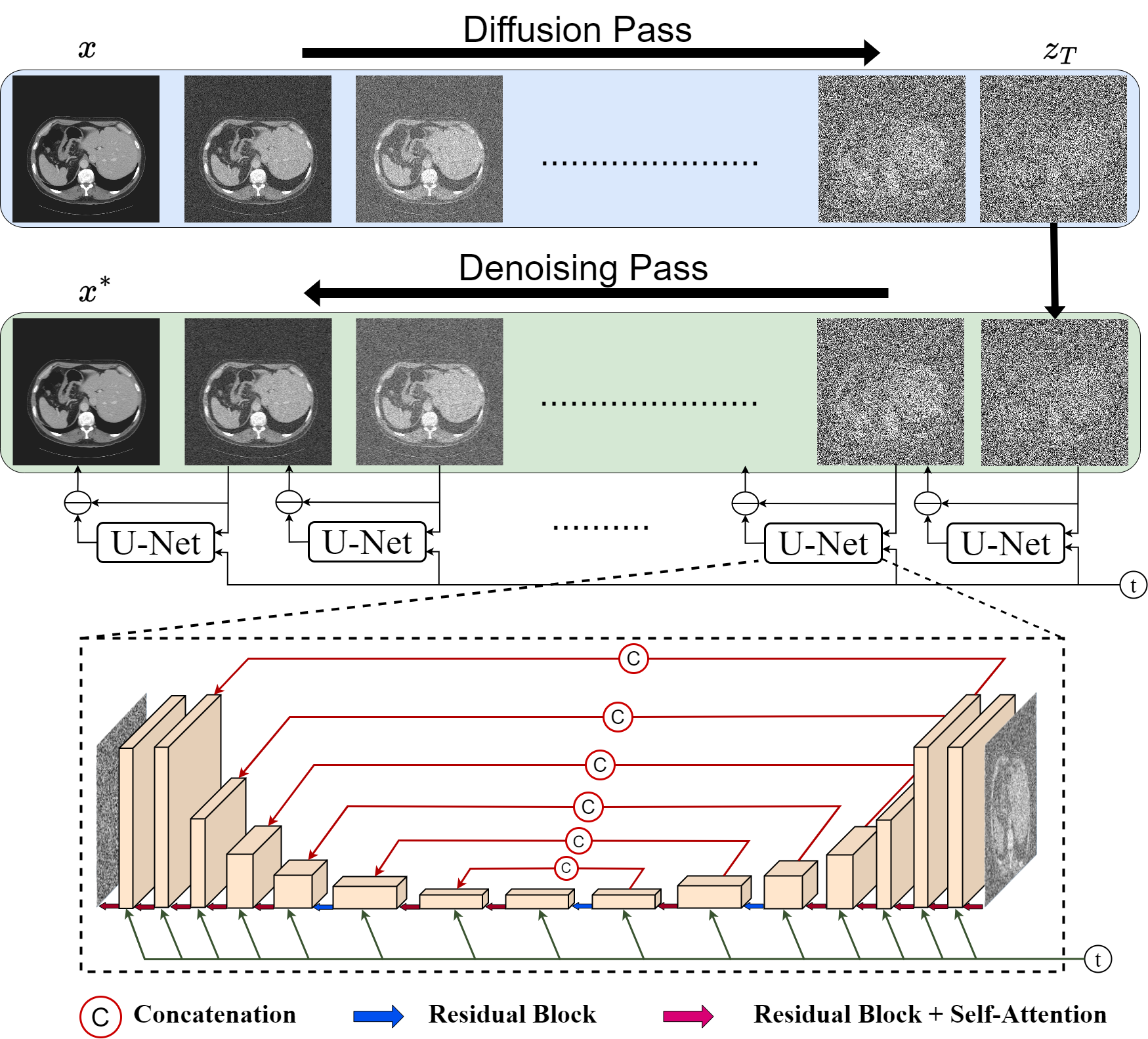}
\caption{General overview of the diffusion model, which consists of two main phases: the diffusion phase, where noise is gradually added, and the denoising phase, where noise is progressively removed using a U-Net-like architecture.}
\label{fig:diff}
\end{figure}

\textbf{Diffusion Models}: Denoising Diffusion Probabilistic Models (DDPMs) have shown great potential in various computer vision tasks. DDPMs consist of two main phases: the forward (or diffusion) process and the reverse (or denoising) process as summarized in Figure \ref{fig:diff} \cite{sohl2015deep, song2020score, ho2020denoising, prince2023understanding}. 

In the forward process, a white noise is gradually added to a real image until it becomes a noisy image with a standard normal distribution. The intermediate latent variables are denoted as \( z_1, z_2, \ldots, z_T \). Specifically, at the first step, the noisy image \( z_1 \) is obtained as:

\begin{equation}
z_1 = \sqrt{1 - \beta_1} \cdot x + \sqrt{\beta_1} \cdot \epsilon_1
\end{equation}

where \( \beta_1 \) is the noise schedule hyperparameter for the first step, \( x \) is the original input image, and \( \epsilon_1 \) is a standard normal noise term. For subsequent steps \( t \), the noisy image \( z_t \) is computed as:

\begin{equation}
z_t = \sqrt{1 - \beta_t} \cdot z_{t-1} + \sqrt{\beta_t} \cdot \epsilon_t \quad \forall \, t \in \{2, \ldots, T\}
\end{equation}

where \( \beta_t \) is the noise schedule at step \( t \), and \( \epsilon_t \) is a standard normal noise term. The hyperparameter \( \beta_t \) determines how quickly the input becomes noisy by attenuating the input data by a factor of \( \sqrt{1 - \beta_t} \) and adding noise by a factor of \( \sqrt{\beta_t} \). For large \( t \), where calculating the intermediate variable is time-consuming, the forward pass can be simplified as:

\begin{equation}
z_t = \sqrt{\alpha_t} \cdot x + \sqrt{1 - \alpha_t} \cdot \epsilon
\end{equation}

where \( \alpha_t = \prod_{s=1}^t (1 - \beta_s) \). 

The goal of the reverse process is to iteratively remove the added noise to recover the original data \( x \). The reconstruction of \( x \) from \( z_t \) can be expressed as:

\begin{equation}
x = \frac{1}{\sqrt{\alpha_t}} z_t + \frac{\sqrt{1 - \alpha_t}}{\sqrt{\alpha_t}} \cdot \epsilon
\end{equation}

To achieve this, a neural network \( g_t[z_t, \phi_t] \) is trained to predict the added noise at step \( t \) and remove it from \( z_t \) to obtain \( z_{t-1} \). The loss function used for training the network at each step \( t \) is:

\begin{equation}
\mathcal{L}_t = \Vert g_t[\sqrt{\alpha_t} \cdot x_t + \sqrt{1 - \alpha_t} \cdot \epsilon, \phi_t] - \epsilon \Vert^2
\end{equation}

During sampling, a final latent variable \( z_T \) is generated from a standard normal distribution, and the noise is gradually removed over \( T \) steps, which is fixed during training. In the reverse process, a single U-Net architecture is used as the denoising network \( g \) for all steps. The timestep information is passed as positional embeddings by concatenating these embeddings with the features at each layer of the U-Net.

Similar to cGANs, conditional information in DDPMs has been extended through various mechanisms, including classifier guidance \cite{dhariwal2021diffusion} and classifier-free guidance \cite{ho2022classifier}. In the classifier guidance approach, an auxiliary classifier is used to guide the denoising process, influencing the generation based on the desired class. On the other hand, classifier-free guidance embeds the class information directly into the model, passing it through all layers of the U-Net architecture, similar to how time step information is incorporated. Additionally, a technique where the conditional information is randomly dropped during training has been employed to enhance the model's robustness.
Another idea that has demonstrated impressive performance involves modifying the U-Net to produce two noise predictions: one conditional and one unconditional \cite{ho2022classifier}. The difference between these predictions represents the class-specific features, effectively enabling the model to distinguish and utilize class information during the denoising process.

\subsection{State-of-the-Art of Generative Models in MIS}
Generative models have been exploited in different manners, including augmenting the training data through a generative model then training a separate model to perform the segmentation using real and generated data \cite{sandfort2019data, mahmood2019deep},  a segmentation model is adversarialy trained to perform the segmentation task \cite{zhang2018translating, huo2018synseg, wu2024medsegdiff, wu2024medsegdiff2}, semi-supervised learning segmentation \cite{zhang2017deep} and self-supervised learning \cite{ma2021self, kim2023diffusion, kim2024c}.

\begin{figure*}
\centering
\includegraphics[width = 7in, height=3in]{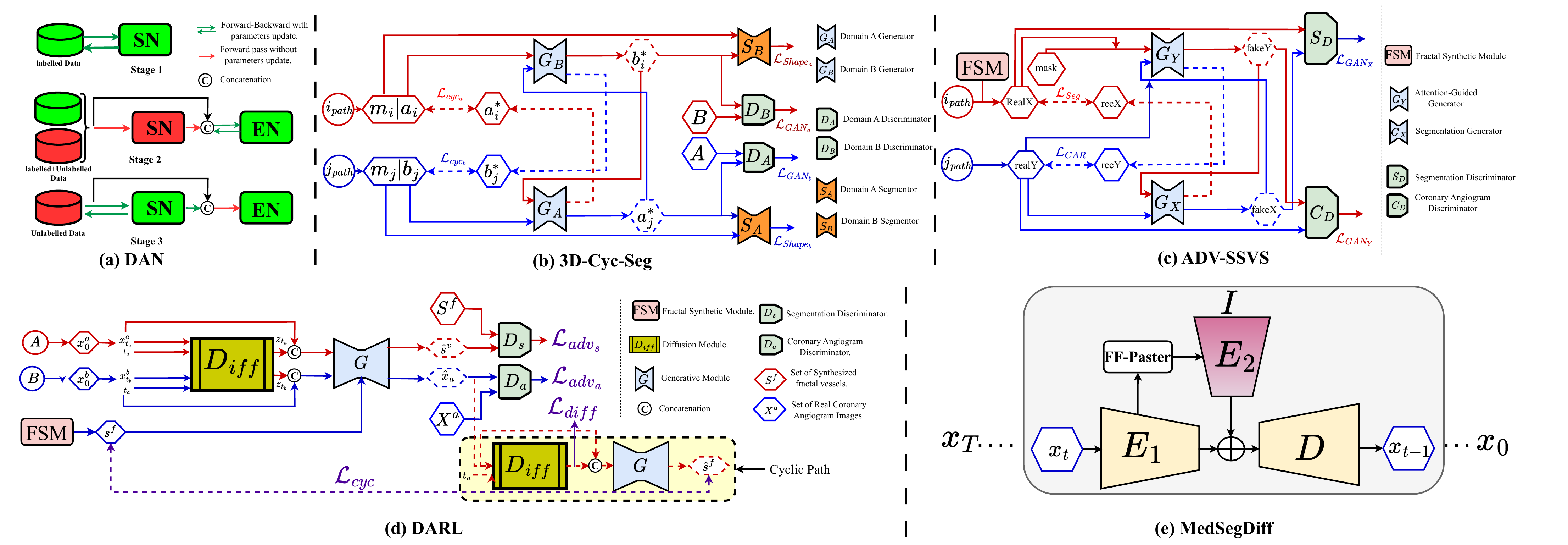}
\caption{Examples of generative models in MIS. a. DAN \cite{zhang2017deep}. b. 3D-Cyc-Seg \cite{zhang2018translating}, $i_{path}$ and $j_{path}$ represent the cyclic translation and segmentation process from domain A to B and from domain B to A, respectively.
c. ADV-SSVS \cite{ma2021self},  X and Y represent the segmentation map and coronary angiogram image, receptively. d. DARL \cite{kim2024c}, a hybrid GAN-Diffusion Model. e. MedSegDiff \cite{wu2024medsegdiff}. Better view in Colors.}
\label{fig:advseg}
\end{figure*}

One of the primary works utilizing adversarial training for medical imaging segmentation is presented in \cite{zhang2017deep}. As shown in Figure \ref{fig:advseg}.a, their approach involves two networks: a segmentation network (SN) and an evaluation network (EN), which leverage both annotated ($D_{a}$) and unannotated data ($D_{u}$). The training process consists of  three stages. In the first stage, the SN is trained on the labeled data $D_{a}$ using a multi-class cross-entropy loss function. In the second stage, the trained SN generates segmentation probability maps for both $D_{a}$ and $D_{u}$. These probability maps are then combined with the input images through element-wise multiplication, resulting in feature maps $f_{ca}$ for $D_{a}$ and $f_{cu}$ for $D_{u}$. Then, EN is trained using $f_{ca}$ and $f_{cu}$, treating them as true and false segmentation predictions, respectively. In the final stage, the SN undergoes adversarial training by incorporating the reverse gradient of $f_{cu}$ to enhance segmentation accuracy for $D_{u}$. This is done gradually to make the EN unable to distinguish between the segmented labeled images and those from the unlabeled set, thereby exploiting the unannotated data to improve segmentation performance.

In \cite{zhang2018translating}, Z. Zhang {\it et al.} leveraged CycleGAN \cite{zhu2017unpaired} for unpaired 3D CT-MRI translation and segmentation as depicted in Figure \ref{fig:advseg}.b. To maintain anatomical consistency during cross-modality translation, they introduced a shape-consistency loss function. This function provides additional supervision by mapping both the translated and original images to a common semantic space via a segmentation network. A segmentation architecture was used after each generator to segment the translated scans, aiming to minimize geometric distortions and complement the adversarial loss, thereby enhancing geometric fidelity during generation. To evaluate the quality of the generated translations, the segmentation architecture was trained using a combination of real and online-generated scans. Results on a 3D cardiovascular dataset demonstrated that augmenting the data with their generated scans improved segmentation performance.

In \cite{huo2018synseg}, a similar approach was proposed, where only the source modality's semantic segmentation labels are available. In their SynSeg-Net architecture, only one segmentation network (Seg) is added after the first generator, which translates the source image to the target domain. The SynSeg-Net, including the Seg architecture, which is trained end-to-end using a weighted loss function comprising two adversarial losses, two cycle-consistency losses, and a segmentation loss. During testing, only the Seg network is used to directly segment real target domain images. Notably, the segmentation labels for the target domain were not available during training, so the Seg architecture was trained using only the source domain's semantic labels. Their approach demonstrated efficiency in two tasks: spleen segmentation in MRI-to-CT synthetic splenomegaly segmentation and brain segmentation in CT-to-MRI synthetic TICV segmentation, outperforming generative methods and achieving results comparable to supervised methods.

Y. Ma \textit{et al.} utilized CycleGAN in a self-supervised manner for coronary vessel segmentation, eliminating the need for labeled data as depicted in Figure \ref{fig:advseg}.c \cite{ma2021self}. They introduced a Fractal Synthetic Module (FSM) to generate vessel segmentation masks (\textit{RealX}). In the first CycleGAN path ($i_{path}$), the synthesized fractal mask (\textit{RealX}) and a mask frame (\textit{frame}), which do not visualize blood vessels, are used by an Attention-Guided Generator ($G_Y$) to create a coronary angiogram (\textit{FakeY}). The second generator ($G_X$), a segmentation generator, then segments \textit{FakeY} into a vessel segmentation mask (\textit{RecX}). In the second path ($j_{path}$), a real coronary angiogram (\textit{RealY}) is processed by the segmentation generator to produce a vessel segmentation mask (\textit{FakeX}), which is then passed to the attention-guided generator alongside \textit{RealY} to reconstruct the input angiogram (\textit{RecY}). A weighted loss function is employed, comprising four components: (i) discriminator loss between \textit{FakeY} and real angiograms ($\mathcal{L}_{GAN_Y}$), (ii) discriminator loss between the segmentation of real angiograms (\textit{FakeX}) and fractal-generated segmentation masks (considered real): $\mathcal{L}_{GAN_X}$, (iii) cycle-consistency loss between \textit{RecX} and \textit{RealX}, which is the segmentation loss $\mathcal{L}_{Seg}$, and (iv) cycle-consistency loss between \textit{RecY} and \textit{RealY} ($\mathcal{L}_{CAR}$). The model leverages adversarial MSE loss, $L_1$ loss, and binary cross-entropy loss for the discriminators, angiogram reconstruction, and segmentation mask reconstruction, respectively. During testing, the coronary angiogram is directly input into the segmentation generator to produce the vessel segmentation mask. The results demonstrated competitive performance compared to supervised approaches, even in domain adaptation scenarios, despite the absence of labeled segmentation masks during training.

\begin{table*}
\small
\centering
\caption{Summary and characteristics of reviewed state-of-the-art generative models in MIS, including: Datasets, Model Types, and Results (Dice Score). 'Dts' Corresponds to Results on the Datasets Listed in the Second Column, Following the Same Dataset Order. 'Transl,' 'Seg,' and 'Aug' Refer to Imaging Modality Translation, Segmentation, and Augmentation, respectively.  }
\label{tab:gen}
\scalebox{0.8}{\begin{tabular}{|p{1.3cm}|p{11.1cm}|p{3cm}|p{5.9cm}|}
\hline
\textbf{Ref} & \textbf{Datasets and Details}& \textbf{Model Type} &\textbf{Results} \\
\hline

2017 \cite{zhang2017deep}& - Gland Segmentation  -Fungus Segmentation &  GAN& -Dts1: 91.6 (Part 1), 85.5 (Part 2)  -Dts2: 93.64\\\hline
\, \newline  2018 \cite{zhang2018translating}& -3D cardiovascular Segmentation on CT-scans \newline -3D cardiovascular Segmentation on MRI scans (Both datasets have five anatomical regions: The endocardium of all four cardiac chambers and the left ventricle epicardium).&   CycleGAN (Transl \& Seg)&  -Dts1: 74.4   \newline -Dts2: 73.2\\\hline

2018 \cite{han2018spine}& -Semantic segmentation of multiple spinal structures, which are intervertebral discs, vertebrae,andneural foramen NFS, IDD, and LVD in MRI scans.& GAN & -Dts1: 87.1 \\\hline

2018 \cite{huo2018synseg}& -Spleen segmentation in MRI-to-CT synthetic splenomegaly segmentation. \newline -Brain segmentation in CT-to-MRI synthetic TICV segmentation.&  CycleGAN (Transl \& Seg)& -Dts1: 89.5  \newline -Dts2: 96.3 \\\hline
2019 \cite{mahmood2019deep}& -Multi-Organ Nuclei
Segmentation in Histopathology Images &  CycleGAN (Aug) \& cGAN (Seg)& -Dts1: 86.6 \\\hline
\, \newline  2019 \cite{sandfort2019data}& -Segmentation of Kidney, Liver, and Spleen in contrast CT-scans dataset (real+synth). \newline-Segmentation of Kidney, Liver, and Spleen in non-contrast CT-scans dataset (real+synth).& CycleGAN (Transl for Aug) then U-Net (Seg)& -Dts1: 93.2  \newline  -Dts2: 74.7\\\hline
2021 \cite{ma2021self}& -Coronary Vessel Segmentation.&  CycleGAN& -Dts1: 55.7\\\hline

\,  \newline \, \newline 2023 \cite{kim2023diffusion} & The model is trained for Vessel Segmentation using  unlabeled X-ray coronary angiography disease (XCAD) dataset \cite{ma2021self}. Then evaluated in the following datasets: \newline -XCAD test set \cite{ma2021self}. -134 XCA dataset \cite{cervantes2019automatic}  -30 XCA dataset \cite{hao2020sequential}. \newline-DRIVE dataset  -STARE dataset \cite{hoover2000locating}.& \, \newline \, \newline \, Hybrid&  \, \newline \,  \newline-Dts1: 63.6. -Dts2: 59.5  -Dts3: 57.2. \newline-Dts4: 52.5  -Dts5: 50.8      \\\hline


\,  \newline 2024 \cite{kim2024c} & Similar evaluation protocol and datasets as in \cite{kim2023diffusion}: \newline -XCAD test set \cite{ma2021self}. -134 XCA dataset \cite{cervantes2019automatic}  -30 XCA dataset \cite{hao2020sequential}. \newline-DRIVE dataset  -STARE dataset \cite{hoover2000locating}.&  \, \newline   Hybrid
& \,   \newline-Dts1: 66.1. -Dts2: 70.0.  -Dts3: 62.1. \newline -Dts4: 51.0.  -Dts5: 52.2.      \\\hline

\, \newline 2024 \cite{wu2024medsegdiff}& -Optic-Cup  (REFUGE-2 dataset \cite{fang2022refuge2}, Fundus) 
\newline -Brain-Turmor Thyroid Nodule (BraTS-2021 dataset \cite{baid2021rsna}, MRI)
\newline -Thyroid Nodule (DDTI dataset \cite{pedraza2015open}, ultrasound).& \, \newline  Conditional DDPM& -Dts1: 86.9   \newline -Dts2: 89.9 \newline -Dts3: 86.1\\\hline
\,  \newline 2024 \cite{wu2024medsegdiff2}& -AMOS22 dataset \cite{ji2022amos}.  -BTCV dataset \cite{gibson2018automatic}. \newline -BraTS-2021 dataset \cite{baid2021rsna}.  -ISIC2018 dataset \cite{codella2019skin}. \newline -DDTI dataset \cite{pedraza2015open} -REFUGE-2 dataset (Disc and Cup) \cite{fang2022refuge2}& 
\, \newline\,  Conditional  DDPM&
-Dts1: 90.1.  -Dts2: 89.5  \newline -Dts3: 90.8.     -Dts4: 93.2   \newline -Dts5: 88.7. -Dts6: (96.7, 87.9)
\\\hline

\end{tabular}}
\end{table*}

In \cite{kim2024c}, B. Kim \textit{et al.} advanced the self-supervised methodology proposed by Y. Ma \textit{et al.} \cite{ma2021self} by introducing a hybrid Diffusion-Adversarial representation model. Their approach leverages background and angiography blood vessel scans, which correspond to retinal scans before and after the injection of a contrast agent, respectively. These images are considered unpaired (non-aligned) due to differences in the time of recording and patient movement. The unpaired inputs are denoted as \( x_0^b \) for background images and \( x_0^a \) for angiograms. The proposed method comprises a diffusion network (\( D_{\text{iff}} \)), a generator (\( G \)), and two discriminators (\( D_s \) and \( D_a \)) as shown in Figure \ref{fig:advseg}.d. The overall approach is divided into two paths, \( A \) and \( B \). In path \( A \), a noisy angiogram \( x_{ta}^a \) is fed into \( D_{\text{iff}} \) to produce latent features \( z_{t_a} \). These features are then concatenated with \( x_{ta}^a \) and passed to the generator \( G \) to generate the corresponding segmentation mask \( \hat{s}^v \). Path \( B \) involves feeding a noisy background image \( x_{ta}^b \) into \( D_{\text{iff}} \) to extract latent features \( z_{t_b} \), which are then combined with the input \( x_{ta}^b \) and passed through the generator \( G \) to generate a synthetic angiogram \( \hat{x}_a \). To ensure the realistic representation of blood vessel structures in \( \hat{x}_a \), fractal vessel-like masks $s^f$ (realX) similar to those in \cite{ma2021self} are injected into the generator layers using switchable SPADE layers \cite{park2019semantic}. Both paths share parameters, with the primary difference being in the generator: during the generation of \( \hat{x}_a \) (path \( B \)), the SPADE layers are active, whereas they are replaced by instance normalization layers in path \( A \).

The discriminators \( D_s \) and \( D_a \) are adversarially trained to distinguish between real and fake segmentation masks (with fractal vessels considered as real segmentation labels) and angiograms, respectively. The model employs a total of five losses: two adversarial losses ($\mathcal{L}_{adv_s}$ and $\mathcal{L}_{adv_a}$) to train the discriminators and generators, a diffusion loss (denoising) applied in path B ($\mathcal{L}_{diff}$), and a cycle loss ($\mathcal{L}_{cyc}$). The cycle loss is crucial as the generated fractal vessel mask using FSM ($s^f$) serves as the segmentation label for the generated angiogram image \( \hat{x}_a \). To enforce consistency, \( \hat{x}_a \) is passed through path \( A \) to obtain the segmentation mask \( \hat{s}^f \), and a cycle reconstruction loss is applied between \( \hat{s}^f \) and \( s^f \). Unlike conventional DDPM, this approach does not require iterative denoising during inference. The generator is already trained to produce segmentation masks from noisy \( x_{ta}^a \) inputs, which enhances its efficiency in handling noisy cases. Comparative experiments demonstrate that this proposed approach significantly outperforms both unsupervised and self-supervised methods on angiography datasets, including X-ray coronary angiography and retinal images.

In \cite{wu2024medsegdiff}, DDPM are exploited with a conditional approach called MedSegDiff, in which the input image is used to guide the denoising process of the noised segmentation mask, as shown in Figure \ref{fig:advseg}.e. To achieve this, a ResUNet architecture with two encoders (\(E_1\) and \(E_2\)) and one decoder (\(D\)) is adopted.
The two encoders have distinct roles: \(E_1\) processes the noised segmentation map at time step \(t\) (\(x_t\)), while \(E_2\) receives the input image \(I\). The input image \(I\) contains full information about the target regions (e.g., infection areas), but these regions are often difficult to distinguish from the background. Conversely, \(x_t\) progressively enhances the target regions through the denoising process. To enable mutual complementarity, segmentation information from \(x_t\) (encoded by \(E_1\)) is integrated into the conditional branch \(E_2\) at different layers via an FF-Parser block and an attention mechanism.
The FF-Parser block applies a Fourier transform to filter out high-frequency noise from the features encoded by \(E_1\). These refined features are then used to guide the features from \(E_2\) through an attention mechanism, ensuring that the segmentation information is accurately aligned with the input image details.
The decoder \(D\) receives the combined output of the final embedded features from both encoders, which are summed before being passed to \(D\). This enables the decoder to reconstruct a refined segmentation mask.
To further enhance the accuracy of the segmentation, the STAPLE ensembling algorithm is applied, which combines multiple segmentation outputs to produce a more robust and precise final segmentation mask.

\subsection{Discussion}
As discussed in the aforementioned state-of-the-art works and illustrated in Table \ref{tab:gen}, generative models are garnering increasing attention in the field of medical imaging segmentation, contributing to performance levels comparable to those of expert radiologists. It is observed that generative models are not limited to traditional data augmentation through the generation of new samples. Their application has expanded to more advanced scenarios, including semi-supervised learning, self-supervised learning, and even the direct training of segmentation networks. Generative models used in medical image segmentation can be categorized into adversarial models (GANs) \cite{zhang2017deep, huo2018synseg, sandfort2019data, ma2021self}, diffusion models \cite{wu2024medsegdiff, wu2024medsegdiff2}, and hybrid models \cite{kim2023diffusion, kim2024c}. Among these, GANs have been the most widely adopted, particularly CycleGAN, which enables the exploitation of different medical imaging modalities through image-to-image translation, providing effective solutions in various medical imaging contexts. In recent years, diffusion models and hybrid approaches have gained significant interest, pushing the boundaries of segmentation performance to new heights. These models have shown great potential in addressing challenges across different medical imaging modalities and tasks. However, one of the limitations in the field is the lack of a unified evaluation protocol and dataset, making it difficult to compare different approaches. Each study often introduces its own evaluation criteria and datasets, which hinders the establishment of a standardized benchmark.


\section{Few-shot Semantic Segmentation (FSS)}
\label{Sec:FFS}

Deep learning architectures have demonstrated significant success in medical imaging segmentation, particularly when large, well-labeled datasets are available. However, creating such datasets is labor-intensive and requires substantial time and expertise from medical professionals like radiologists. Furthermore, models trained on these datasets often struggle to generalize to unseen classes. The use of Few-Shot Semantic Segmentation has shown a potential to cope with these challenges by learning new classes from only a few labeled examples. 


\subsection{Problem Formulation}

In typical image segmentation paradigm, the model $f$ is trained with a pair of \texttt{<image-mask>} ($D_{train} = \{(I_{tr}^i, Y_{tr}^i)\}_{i=1}^{N_{train}}$) and tested on pair of \texttt{<image-mask>} ($D_{test} = \{(I_{ts}^i, Y_{ts}^i)\}_{i=1}^{N_{test}}$), where $N_{train}$ and $N_{test}$ are the number of training and testing \texttt{<image-mask>}, respectively. Both training and testing masks share the same classes ($C_{tr}, C_{ts} \in C_l^{H \times W}$), where each pixel in $H \times W$ is assigned to one of $l$ classes. In  contrast, in few-shot segmentation learning paradigm $C_{tr}^{H \times W} \cap C_{ts}^{H \times W} = \emptyset$. Also, $C_{tr}$ and $ C_{ts}$ are called $C_{seen}$ and $ C_{unseen}$, respectively, because the new classes to be predicted are unseen during the training phase. The aim of few-shot learning is to learn new semantic classes segmentation from few samples $k$ (1 to 5) without the need to retrain the model for the new classes.
In the inference phase, the model $f$ exploits a small set of $k$  \texttt{<image-mask>} pairs $S = \{(I_{s}^i, Y_{s}^i)\}_{i=1}^k$ called support set to predict $\hat{M}$ of the query image $I_q$, where $Y_{s}^i \in C_{unseen}$: $\hat{M} = f (I_q; S)$. 

To match the above scenario, train and test splits ($D_{train}, D_{test}$) are constructed from \texttt{<image-mask>} pairs that have two non-overlapping classes sets ($C_{seen}, C_{unseen}$), respectively. The two splits are formed as episodes, where each episode composes of supporting set $S$ and a query image $I_q$. Hence,  ($D_{train} = \{(S^i,  I_q^i)\}_{i=1}^{N_{train}}$) and ($D_{test} = \{(S^i, I_q^i)\}_{i=1}^{N_{test}}$), with $N_{train}$ and $N_{test}$ are the number of episodes for training and testing, respectively. The training support of one episode consists of pair of \texttt{<image-mask>}, where the mask contains only the semantic classes from $C_{seen}$. Similarly, the testing support of one episode consists of pair of \textless image-mask \textgreater, where the mask contains only the semantic classes from $C_{unseen}$. This ensures that the model has not seen these classes during the training and learn to predict their semantic segmentation mask using only the few-shot samples having $C_{unseen}$ classes \cite{shaban2017one, wang2019panet}. 

In summary, during training, the model $f$ learns from the episodes of $D_{train}$ to extract class-specific features from the annotated support set. It then segments the query images by utilizing the distilled knowledge from these support images. During testing, the model extracts features from the testing support images, which are labeled with classes unseen during training. Consequently, the model can segment novel classes in the testing query set $Q$. It should be noted that when only one support image is available in few-shot learning, this called one-shot learning. Thus, one-shot semantic segmentation is abbreviated as OSS.


\subsection{Used Datasets in FSS for MIS}
\label{fsdts}

In order to match the few-shot segmentation scenarios, many medical imaging segmentation datasets were adopted to create episodic training, testing scenario by the state-of-the-art approaches. These datasets includes Visceral dataset \cite{jimenez2016cloud}, Abd-CT (Synapse) \cite{landman2015miccai}, Abd-MRI (CHAOS) \cite{kavur2021chaos}, Cardiac-MRI \cite{zhuang2018multivariate}, ABD-110 \cite{tang2021spatial}, and Prostate-MRI \cite{li2023prototypical}.

One of the primary works for few-shot volumetric medical imaging segmentation, A. G. Roy \textit{et al.} used Visceral dataset \cite{jimenez2016cloud} to construct volumetric segmentation evaluation scenario \cite{roy2020squeeze}. Visceral dataset consists of two splits silver corpus (with 65 scans) and gold corpus (20 scans), each contains four classes: Liver, Spleen, R/L Kidney and R/L Psoas Muscle. In \cite{roy2020squeeze}, the silver and gold corpus are used as training, and testing splits, respectively. The testing split is formulated as follows: 14 validation scans, 1 support scan, and 5 testing scans.
The evaluation is performed in four trials, where in each trail one class is select as unseen class $c_{un}$ (testing class), and the remaining three classes are considered as the training classes ($C_{seen}$). During the training, mini-batch is constructed using a Batch Sampler, which selects randomly one class $c$ from the training classes $C_{seen}$, then two slices are randomly selected from the training data to constructed episodic $(S, Q)$ sample, where $S=(I_s, Y_s)$ and $Q=(I_q, Y_q)$. Finally, both supporting and query masks are binarized by considering the query class $c \in C_{seen}$ as foreground and the rest as background. 

In the evaluation phase, for the given testing class $c_{un_i}$ of trial $i$, where $i =1,..,4$, the range of slices where the class appears is identified as $[S^s, S^e]$ and $[Q^s, Q^e]$ for the support scan and query scans, respectively, where $s$ and $e$ indicating the starting and end indexes. In order to simulate the few-shots scenario evaluation with only $k$ slices annotation budget, the intervals $[S^s, S^e]$ and $[Q^s, Q^e]$ are splitted into $k$ groups of slices. In order to ensure that the query image and support image have minimal dissimilarity, the query slices for each group uses the center slice of the corresponding group from the support scan. This evaluation scenario ensures practical efficiency with only $k$ annotation budget. We will refer to this evaluation scenario using Visceral dataset as FSS-Visceral.

In \cite{ouyang2020self}, C. Ouyang \textit{et al.} expanded the evaluation framework of the previous approach to encompass a broader range of datasets: three in total (Abd-CT, Abd-MRI, and Cardiac-MRI). The Abd-CT and Abd-MRI datasets include the classes left kidney, right kidney, spleen, and liver, while the Cardiac-MRI dataset includes the classes left-ventricle blood pool (LV-BP), left-ventricle myocardium (LV-MYO), and right ventricle (RV). They conducted evaluations using a five-fold cross-validation methodology with two distinct settings.

In this five-fold cross-validation scenario, each fold served as a testing fold, and four trial experiments were executed. In each trial, one class was designated as the testing class, and the data from the remaining four folds were used for training. The first evaluation setting mirrored the approach in \cite{roy2020squeeze}, where testing classes might appear in the training data as background. Conversely, in the second setting, they excluded slices containing the testing classes from the training data, ensuring these classes were absent in any context (including as background). This methodology was specifically applied to the Abd-CT and Abd-MRI datasets.

Similar to the methodology in \cite{roy2020squeeze}, they restricted the annotation budget to $k$, conducted grouping, and matched query and support similarities during the testing phase, with $k$ set to 3. These two evaluation settings have since been adopted by numerous state-of-the-art studies \cite{ouyang2020self, tang2021recurrent, lin2023few, zhang2024prototype}. We will refer to the datasets under the first and second evaluation settings for Abd-CT and Abd-MRI as St1-Abd-CT, St1-Abd-MRI, St2-Abd-CT, and St2-Abd-MRI, respectively. Given that Cardiac-MRI was evaluated using only one setting, we will refer to it as St-Cardiac-MRI.

These evaluation settings have been adopted by several contemporary works, including \cite{zhang2024prototype}. In \cite{zhang2024prototype}, the second evaluation setting is applied to the Abd-CT, Abd-MRI, and ABD-110 \cite{tang2021spatial} datasets, which all contain the same classes (left kidney, right kidney, spleen, and liver). Similarly, we will refer to the second evaluation setting for ABD-110 as St2-ABD-110.

Similar evaluation protocols have been employed in state-of-the-art works. In \cite{tang2021recurrent}, five-fold cross-validation using setting 2 is utilized, akin to \cite{ouyang2020self}. For each query image, one random support image is selected, and this process is repeated five times, with the average result considered, as in \cite{wang2019panet}.

In recent works such as \cite{hansen2022anomaly}, the authors argue that limiting the evaluation to only the slices that contains the query class in the 3D scans does not much the real scenario in medical imaging and requires extra supervision and since their model is 3D model. Thus, they evaluated the full 3D scans using the first setting as in \cite{ouyang2020self}. We will refer to this evaluation setting as St1-Abd-MRI* and St-Cardiac-MRI* for Abd-MRI, and Cardiac-MRI datasets, respectively.

Recently, Y. Li \textit{et al.} introduced a new dataset named Prostate-MRI, comprising 589 T2-weighted images collected from seven previous studies, with semantic segmentation labels for eight lower-pelvic structures: bladder, bone, central gland, neurovascular bundle, obturator internus, rectum, seminal vesicle, and transition zone \cite{li2023prototypical}. These classes were divided into four groups, and the experiment includes four trials where, in each trial, one group of classes was considered as the test classes (unseen), while the remaining three groups were considered as the training classes (seen). To ensure a cross-institution experiment, one institution was selected as the unseen institution, specifically institutions 3 or 4. The data from the remaining six institutions were split into training and testing sets. The model was trained using the training data from these six institutions (with seen classes), and all images from the selected testing institution were used as query images. Support images were drawn either from the testing institution or from the testing data of the training institutions \cite{li2023prototypical}. We will refer to this dataset and evaluation protocol as CIFSS-Prostate-MRI.


\subsection{State-of-the-Art of FSS in MIS}
The main challenges in FSS from medical imaging are accurately delineating the segmentation foreground boundaries from the background. The background is usually inhomogeneous, large, and contains tissues similar to the foreground, which exacerbates another significant challenge: class imbalance. Table \ref{tab:few} summarizes state-of-the-art approaches for few-shot semantic segmentation of medical imaging. In fact, FSS in medical imaging ranges from the traditional conditional approach \cite{roy2020squeeze}, to the recent prototypical networks \cite{ouyang2020self,tang2021recurrent, hansen2022anomaly,lin2023few,huang2023rethinking, zhang2024prototype} and hybrid approaches \cite{wu2022dual, li2023prototypical, feng2023learning} 
this last exploits concepts from both conditional and prototypical networks.

\begin{figure}[htp]
\centering
\includegraphics[width = 3.5in, height=2.5in]{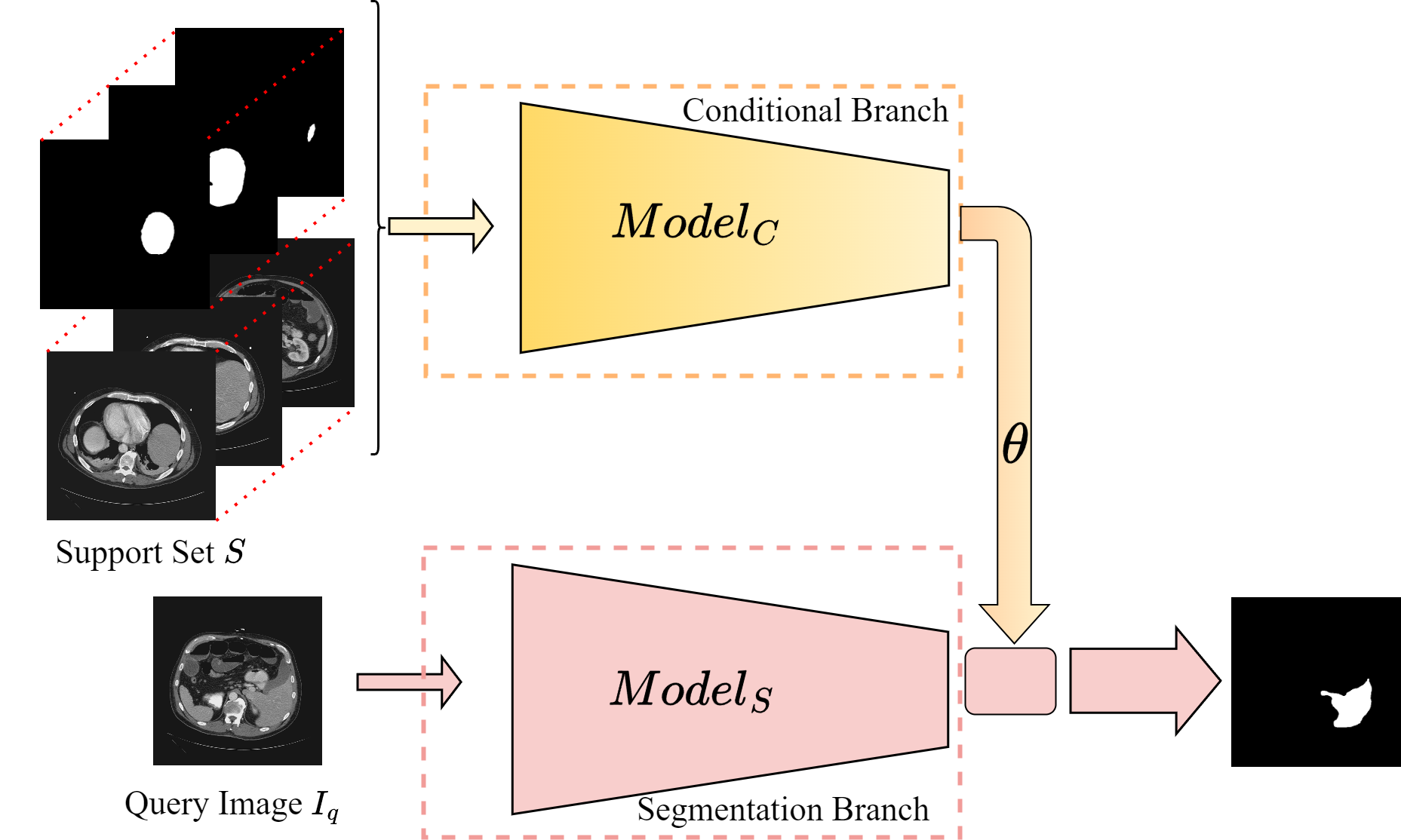}
\caption{Conditional FSS overview: Conditional Branch for Parameter Generation and Segmentation Branch for Query Segmentation.}
\label{fig:cffs}
\end{figure}
\begin{figure}[htp]
\centering
\includegraphics[width = 3.5in, height=2.5in]{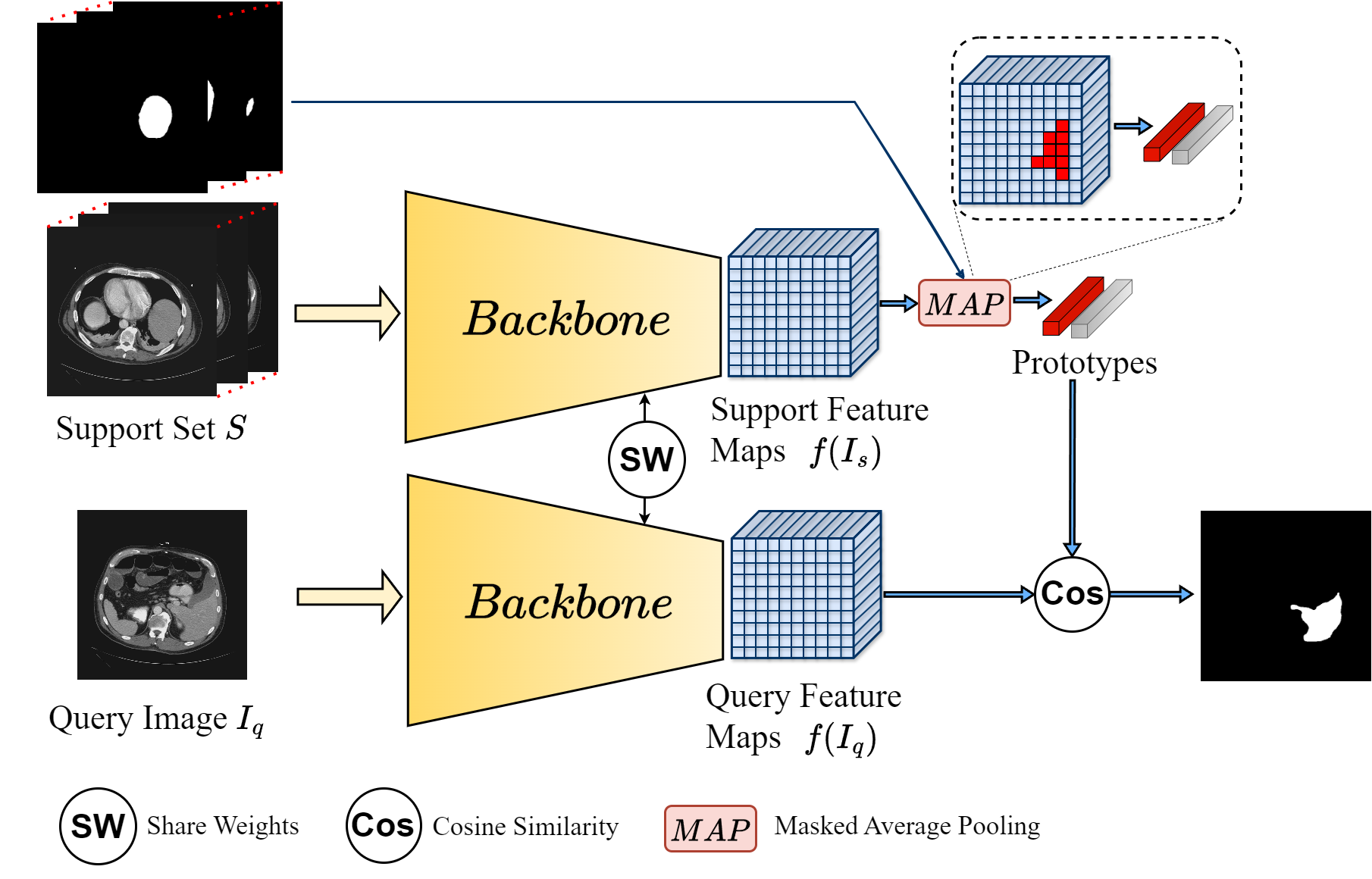}
\caption{Prototype FSS overview: Shared Backbone, Prototype Generation, and Similarity-Based Segmentation}
\label{fig:pffs}
\end{figure}

\textbf{Conditional Approach Definition}: As summarized in Figure \ref{fig:cffs}, Conditional FSS approach uses one deep architecture to generate a set of parameters $\theta$ from the support (\texttt{<image-mask>}), which called conditional branch. The generated parameters $\theta$ are used to tune the segmentation process on the extracted features from the query image (usually different architecture), and this is called segmentation branch.  

\textbf{Prototypical Approach Definition}: In contrast to Conditional approach, Prototypical Networks used for FSS exploits one backbone (usually VGG or Resnet) to embed the support and query images (the episode images) into embedding features space as depicted in Figure \ref{fig:pffs}. Using Masked Average Pooling (MAP) on the support set produces classes prototypes. The query image embedded features are used to classify each pixel of the input query image into the class of the nearest class prototype obtained from the support set. This is performed by using a distance function and softmax to obtain the classes probabilities mask, where cosine and squared Euclidean have been widely used as distance functions \cite{wang2019panet}. In the training phase, the segmentation loss is calculated between the query and the predicted probabilities masks.

In \cite{roy2020squeeze}, A. G. Roy \textit{et al.} proposed conditional one-shot volumetric segmentation approach using vused Visceral dataset \cite{jimenez2016cloud}. Their approach consists of conditional branch, segmentation branch and interaction blocks. In more details, both conditional and segmentation branches are encoder-decoder architectures (U-net like architecture without skip connection). In this work, the author addressed one limitation of the state-of-the-art approaches, which mainly adopts pretrained models for conditioner and segmentor branches and this required only one connection level, which is usually at the end. In contrast, in this work, they proposed stronger interaction block using \textit{Channel Squeeze \& Spatial Excitation (sSE)} at multiple encoder-decoder levels of the conditioner and segmentor branches. Between each of the same level features of the encoder, bottleneck and decoder of both conditioner and segmentor, sSE block is a spatial attention that exploits the conditioner features through squeezing operation to perform excitation (re-calibration) on the segmentor feature maps. The squeezing operation is 1 by 1  convolutional layer that transforms the conditioner feature maps channels' number to 1 then followed by  sigmoid layer to rescale activations to $[0, 1]$ and the excitation is performed by multiplying the obtained weights by the segmentor feature maps.

In \cite{ouyang2020self}, C. Ouyang \textit{et al.} introduced the SSL-ALPNet framework, based on PANet \cite{wang2019panet}, to facilitate training without manual annotations by using superpixels to generate pseudo-labels. This method enables learning from unlabeled images, enhancing the diversity and generalizability of image representations. The process involves generating superpixel pseudo-labels and iteratively selecting one to form a support \texttt{<image-mask>} pair. Data augmentations are applied to generate corresponding query \texttt{<image-mask>} pairs. Besides the masked average pooling, they introduced the Adaptive Local Prototype Pooling (ALP) module, which computes local prototypes by averaging features within a window of size $(L_H, L_W)$. The ensembling of MAP and ALP prototypes, along with the cosine similarity function, are then used to classify the pixels of the query image.

To address the challenges of foreground boundary segmentation and class imbalance in medical imaging segmentation, H. Tang \textit{et al.} proposed a prototype-based approach with two key contributions \cite{tang2021recurrent}. First, they introduced the Context Relation Encoder (CRE), which encodes the context relationship between the foreground and background by calculating the correlation between their masked feature maps. These correlated features are concatenated with features extracted by the backbone network and passed through a 1 by 1 convolutional layer to obtain fused context features and context relation features, referred to as enhanced features. This process is applied to both support and query images, using the union of the support masks for the query image due to the absence of its mask.
The enhanced support features are then used to identify class prototypes via masked average pooling, and the query mask is predicted using enhanced query features and cosine similarity with the support prototypes. Through the second contribution, recurrent mask refinement, the predicted query mask is iteratively refined by using the previous predicted query mask to compute context interaction features and the enhanced query features (instead of using the union of the support masks). A new query mask is then predicted based on these updated enhanced query features, support prototypes, and cosine similarity. This process is repeated with parameter sharing to obtain the final refined query prediction.

To address the challenges of generalization bias towards the training classes and class imbalance in medical imaging, Y. Feng \textit{et al.} proposed a two-branch approach based on prior knowledge, prototype, and attention mechanisms \cite{feng2023learning}. Their method consists of the spatial and segmentation branches, with the spatial branch following a U-Net-like architecture and the segmentation branch employing an encoder-decoder structure. The goal of the spatial branch is to identify the target class location in the query image as prior knowledge, based on the observation that the positions of human organs are consistent. This is achieved by registering the support image to the query image. Subsequently, the support mask is warped using the same registration spatial transformation to obtain the prior query mask.
In the segmentation branch encoder, multi-scale prototype features are extracted using the support and query image-mask pairs, with the query prior mask from the spatial branch being considered as the query mask during both training and testing phases. The similarity between the query maps and the prototype is then calculated. At the final stage, an attention-based fusion module is employed to incorporate the similarity prototype and query prior mask from the spatial branch with the segmentor decoder features at multiple scales, resulting in an accurate segmentation mask.

In \cite{hansen2022anomaly}, S. Hansen \textit{et al.} extended the self-supervised learning approach with superpixels from \cite{ouyang2020self} to supervoxels, leveraging the 3D structure in medical imaging to create uniform pseudo-labels across slices. This method aims to bridge the gap between pseudo-labels used during training and the real unseen labels encountered during testing. To address the challenge of insufficient prototypes due to few support samples with large, heterogeneous backgrounds and relatively small, homogeneous foregrounds, they focused solely on the foreground prototype, drawing inspiration from the anomaly detection literature. Query pixels are classified based on their similarity to the foreground prototype with a learned threshold.

\begin{table*}
\small
\centering
\caption{Summary and characteristics of FSS in MIS. In-house Ultrasound Echocardiography dataset* \cite{feng2023learning} consists of videos of apical 2-, 3-, and 4-chamber views from 60 patients. Left atrium (LA) and Ventricle (LV). The dataset is splitted into 2-fold, for each fold three views of the LA (2C, 3C, 4C) or three views of the LV (2C, 3C, 4C) are considered as the unseen class once per-fold.  }
\label{tab:few}
\scalebox{0.9}{
\begin{tabular}{|p{1.3cm}|p{7.8cm}|p{4cm}|p{5cm}|}
\hline
\textbf{Ref} & \textbf{Datasets, Classes and Evaluation Scenario}& \textbf{Model Type (Backbone)} &\textbf{Results} \\
\hline

2020 \cite{roy2020squeeze} & -FSS-Visceral \cite{roy2020squeeze}.&  Conditional (Unet*, Unet*)& -Dts1: 48.5\\\hline

\vspace{6pt} 2020 \cite{ouyang2020self} & -Abd-CT (St1-Abd-CT and St2-Abd-CT) \newline -Abd-MRI (St1-Abd-MRI and St2-Abd-MRI) \newline - St-Cardiac-MRI &  \vspace{6pt} Prototype (ResNet-101)
& -Dts1: 73.35, 63.02 \newline -Dts2: 73.02, 78.84 \newline -Dts3: 76.90 \\\hline

 2021 \cite{tang2021recurrent}  & -St2-Abd-CT  -St2-Abd-MRI  -St2-ABD-110 &   Prototype (Unet**) & -Dts1: 72.48  -Dts2: 79.26  -Dts3: 81.91 \\\hline

2022 \cite{wu2022dual} &-St1-Abd-CT. -St1-Abd-MRI.  & Hybrid  &  -Dts1: 76.36  -Dts2: 68.16 \\\hline


2022 \cite{hansen2022anomaly}&-St1-Abd-MRI*  -St-Cardiac-MRI*. & Prototype (3D ResNeXt-101)  &  -Dts1: 72.41  -Dts2: 69.62\\\hline

 2023 \cite{lin2023few} & -St2-Abd-CT   -St2-Abd-MRI   -St-Cardiac-MRI&  Prototype (ResNet-50)& -Dts1: 66.59   -Dts2: 75.18  -Dts3: 79.03\\\hline

2023 \cite{li2023prototypical} &- CIFSS-Prostate-MRI \cite{li2023prototypical}& Hybrid (3D UNet) & - Insti 3: 53.51  - Insti 4: 42.85\\\hline

2023 \cite{feng2023learning} & - In-house Ultrasound Echocardiography* \cite{feng2023learning}. 
\newline - St1-Abd-MRI \cite{kavur2021chaos} & Hybrid \newline (Unet, Encoder-Decoder)  &
-Dts1 ( 1-, 5-shots):  87.06, 87.87 \newline -Dts2 ( 1-, 5-shots):  86.37, 88.02
\\\hline

2023 \cite{huang2023rethinking}  & -St1-Abd-MRI   -St-Cardiac-MRI \cite{zhuang2018multivariate} -CIFSS-Prostate-MRI.&   Prototype (ResNet-101)  &-Dts1: 84.56  -Dts2: 84.86  -Dts3: 63.22 
\\\hline

\vspace{6pt} 2024 \cite{zhang2024prototype} & -Abd-CT (St1-Abd-CT and St2-Abd-CT) \newline -Abd-MRI (St1-Abd-MRI and St2-Abd-MRI) \newline -St-Cardiac-MRI. -CIFSS-Prostate-MRI. &  \vspace{6pt} Prototype (ResNet-101)&  -Dts1: 77.23, 74.33 \newline -Dts2: 87.37, 81.53 \newline -Dts3: 87.07. -Dts4: 65.54 \\\hline

\end{tabular}}
\end{table*}


\subsection{Discussion}
As shown in Table \ref{tab:few} and in previous detailed works, few-shot learning for medical imaging segmentation is gaining increasing interest, and the performance gap with supervised scenarios is becoming smaller. In terms of studied tasks, abdominal organ segmentation and myocardial segmentation have been widely investigated as benchmarks. Recently, cross-institution few-shot segmentation was introduced \cite{li2023prototypical}, opening the door to more real-world applications.
Regarding methods, prototype networks have been extensively used. Hybrid approaches have also demonstrated promising performance. ResNets and encoder-decoder architectures (such as U-Net) are commonly employed as backbones for feature extraction. Recently, 3D models \cite{hansen2022anomaly, li2023prototypical} have garnered increasing interest for exploiting contextual relationships between slices for few-shot segmentation, potentially enhancing performance. Many proposed techniques for FSS in medical imaging have shown their efficiency and have been adopted by numerous state-of-the-art approaches. For instance, self-supervised superpixel methods instead of training labels, initially proposed in \cite{ouyang2020self}, were followed by works like \cite{wu2022dual, hansen2022anomaly, huang2023rethinking, zhang2024prototype}. Recurrent Prediction Refinement, proposed in \cite{tang2021recurrent}, has been utilized in various approaches to boost performance, such as \cite{lin2023few}.

Despite these considerable advancements, several aspects require further investigation. Most existing approaches have been evaluated on only a limited number of classes (2-4), primarily focusing on abdominal organ segmentation, indicating a lack of variety in medical imaging tasks. Additionally, evaluation protocols on the same dataset can vary between studies, leading to inconsistencies that make it difficult to compare different methods. The absence of a unified evaluation protocol and the lack of released implementation codes for some notable works further hinder progress in this field. Consequently, each new study must reproduce the experiments of existing state-of-the-art methods for comparison, which is a tedious and time-consuming task. This issue has led to contrasting results for the same approach on the same dataset, further complicating the comparison of different methodologies. Moreover, the performance gap between few-shot learning and fully supervised methods remains significant. For example, in \cite{roy2020squeeze}, the dice score difference ranges from 20-40\% in FSS-Visceral, and in \cite{li2023prototypical}, the dice score difference ranges from 30-40\% in CIFSS-Prostate-MR. In \cite{ouyang2020self}, the difference in dice score is 22\% and 16\% for St1-Abd-CT and St1-Abd-MRI, respectively.

\section{Foundation Models}
\label{Sec:FM}
Inspired by the success of large language models in natural language processing \cite{brown2020language, achiam2023gpt, chowdhery2023palm, touvron2023llama}, similar models have been extended to vision tasks such as CLIP \cite{radford2021learning}, BLIP \cite{li2022blip}. Recently, foundation models for image segmentation have gained substantial interest in the computer vision research community. In particular, the release of models such as Segment Anything Model (SAM) \cite{kirillov2023segment} and Segment Everything Everywhere All at Once (SEEM) \cite{zou2024segment} has sparked discussions about the development of general-purpose segmentation models, both for natural and medical imaging. Despite SAM’s promising performance in natural image segmentation, the applicability of foundation models for MIS remains under investigation due to critical differences in structural complexity, contrast, and inter-observer variability between natural and medical images. To address these challenges, various approaches have been proposed to leverage foundation models in MIS, generally falling into one or multiple of the following categories:  (i) Zero-Shot SAM, (ii) Fine-tuning SAM, (iii) Refined SAM for MIS, (iv) 3D SAM, (v) SAM in semi-supervised learning, and (vi) Automatic prompting SAM, as summarized in Table \ref{tab:found}. The remainder of this section begins with a brief overview of the SAM model, followed by an in-depth discussion of its applications and adaptations across the aforementioned categories in MIS.

\subsection{SAM Background}

\begin{figure}[htp]
\centering
\includegraphics[width = 5.5in, height=2in]{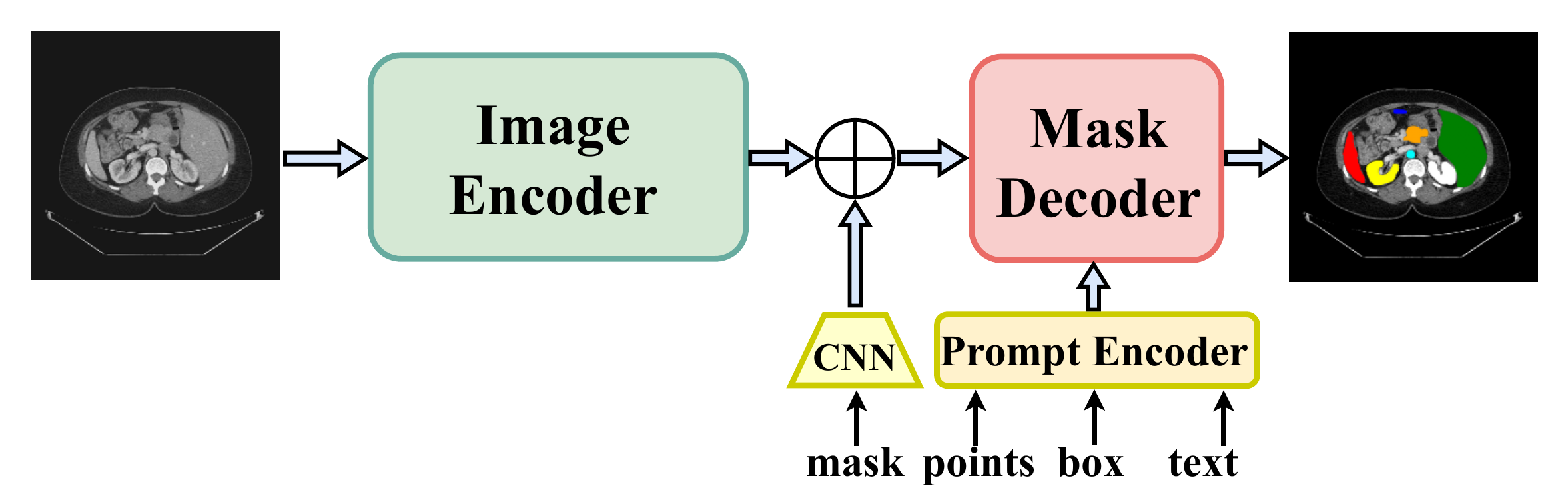}
\caption{General Overview of SAM: Image Encoder, Prompt Encoder, and Mask Decoder for Versatile Mask Generation.}
\label{fig:sam}
\end{figure}

The Segmenting Anything Model was trained on a large natural image segmentation dataset consisting of 1 billion masks and 11 million images \cite{kirillov2023segment}. The aim of SAM is to train a model capable of providing a valid mask using any type of prompt. As shown in Figure \ref{fig:sam}, SAM consists of three main components: (i) Image Encoder, (ii) Prompt Encoder, and (iii) Mask Decoder.

The \textbf{Image Encoder} is a ViT architecture \cite{dosovitskiy2020image} pre-trained on a large dataset using a Masked Autoencoder (MAE) \cite{he2022masked} to embed the input image into high-level features. The \textbf{Prompt Encoder} processes sparse prompts (points, boxes, text) and dense prompts (masks). Points and boxes are embedded using positional encoding \cite{tancik2020fourier}, text prompt is embedded with CLIP's text encoder \cite{radford2021learning}, and dense mask prompt is embedded using a CNN. The \textbf{Mask Decoder} plays a critical role in obtaining the segmentation mask by combining the image embedding and the prompt embeddings.
First, the image and mask embeddings are added element-wise. Inspired by the class token, the output tokens are concatenated with the prompt embeddings and passed through self-attention to make the prompt embeddings more contextually aware of one another. The updated prompt embeddings and image-mask embeddings are iteratively refined through cross-attention in two consecutive layers. Finally, convolutional upsampling is applied to obtain the segmentation mask.

\begin{table*}
\small
\centering
\caption{Summary of foundation models for MIS. FM refers to foundation models, designed for segmenting both seen and unseen medical imaging modalities and targets. TS refers to task-specific models, where foundation models are trained and evaluated exclusively on seen modalities and targets. Bbox refers to bounding box prompting mode.}
\label{tab:found}
\scalebox{0.8}{\begin{tabular}{|p{1.3cm}|p{1.9cm}|p{2.1cm}|p{5cm}|p{9.8cm}|}
\hline
\textbf{Ref}& \textbf{Model Type}&\textbf{Prompt Mode} & \textbf{Dataset \& Imaging Modality}  &\textbf{Results} \\
\hline


2023 \cite{wald2023sam}& Zero-shot (FM)& -Point \newline -Bbox & Abdominal Multi-Organ from  CT-scans (AMOS22 dataset \cite{ji2022amos})& Zero-shot experiments showed that bounding box achieves better performance than point prompt and competitive with supervised. \\\hline

2023 \cite{mazurowski2023segment}& Zero-shot (FM) & -Points. \newline -Bbox. &19 medical imaging datasets& Zero-shot experiments showed that bounding box achieves better performance than point prompt and competitive with supervised. \\\hline

2024 \cite{huang2024segment}& Zero-shot (FM) & -Points - \newline -Bbox \newline -Auto Everything &52 open source datasets, and built a large medical segmentation dataset with 16 modalities, 68 objects, and 553K slice.  & -Manual prompts achieved better performance than Automatic Everything. \newline -Excellent performance for some specific objects and modalities while underperforming for others.  \\\hline \hline

2023 \cite{ma2024segment}& Full Fine-Tuning SAM (FM)&  -Bbox. &More than 1.5 million medical image-mask pairs from 11 modalities for 60 segmentation tasks & -Robust in segmenting Unseen classes and modalities. \newline -Facilitating MIS Annotation.  \newline -Difficulty in segmenting underrepresented modalities and structures with weak boundaries (e.g., vessels). \\\hline

2023 \cite{zhang2023customized} &PEFT SAM (TS)&  Sparse and Dense prompts &Synapse \cite{chen2021transunet} & -The effectiveness of the transferability of SAM through fewer parameters training. \\\hline

2023 \cite{cheng2023Sammed2d} & PEFT SAM (FM)& -Point \newline -Bbox \newline -Mask& 4.6M images and 19.7M masks sourced from public and private datasets, covering 31 major organs and their corresponding anatomical structures across 10 medical imaging modalities. &  -SAM-Med2D excels in segmenting complex organ structures, lesions, and areas with unclear boundaries.
\newline -Compared to SAM, SAM-Med2D performs well across various prompt modes, enabling accurate segmentation in diverse medical imaging scenarios.
\newline -SAM-Med2D effectively handles unseen medical image data, delivering reliable segmentation results.\\\hline

2024 \cite{paranjape2024s}& PEFT SAM (TS) & -Text.& 5 datasets each from different Medical imaging modality.& -Reducing the trainable parameters to only 4\%. \newline -Omitting the need for expert intervention to perform model prompting. \newline -Outperforming SOTA segmentation methods and other SAM-based models.\\\hline

2024 \cite{paranjape2024adaptivesam} & PEFT SAM (TS) & -Text. &5 MIS datasets including surgical, ultrasound, and X-ray modality & \textit{AdaptiveSAM} outperformed zero-shot SAM and other SOTA approaches across surgical, ultrasound, and X-ray modality datasets. \\\hline \hline


2023 \cite{zhang2023sam}& Modified SAM (TS)& -Class trainable prompt.& BCSS \cite{amgad2019structured} and CRAG \cite{graham2019mild}& Experimental results demonstrated the effectiveness of combining histopathology and SAM encoders with automatic prompting. \\\hline

2023 \cite{zhang2023input} & Modified SAM (TS)& Auto prompting. & Three MIS datasets.& Experimental results showed that employing SAM for input augmentation significantly improved the performance.  \\\hline \hline


2023 \cite{gong20233dsam}& 3D SAM (TS) & -Points. &Four public datasets for volumetric tumor segmentation. & -\textit{3DSAM-adapter} outperformed MIS SOTA approaches  with only using one point prompt on 3 out of 4 tasks.   \newline -Performance improved when using more points for prompting (3 and 10 points). \\\hline

2023 \cite{wu2023medical}& 3D SAM (TS) & -Points. \newline -Bbox. & 6 medical image segmentation datasets and 5 medical image modalities&  \textit{Med-SA} Achieved the best performance compared to MIS SOTA methods and foundation models (SAM and MedSAM). \newline -Box prompts with 0.75 overlap outperforming other prompts on most evaluated datasets.  \\\hline

2024 \cite{wang2024sammed3d}& 3D SAM (FM)&  -Points. &  70 public and 24 private datasets consisting of 22K 3D images and 143K corresponding 3D masks across 28 MI modalities. & -Training SAM-Med3D from scratch on a large 3D MI dataset with minimal prompts demonstrated superior performance over 2D SAM adaptations on both seen and unseen targets. \newline -\textit{SAM-Med3D} achieved competitive performance compared with task-specific models using only 1 prompt point and outperformed them in most evaluation datasets with 10 prompt points per volume, excelling on unseen data sources and modalities.\\\hline

2024 \cite{chen2024ma}& 3D SAM (TS)& No prompt&four medical image segmentation tasks, by using 10 public datasets across CT, MRI, and surgical video data.  &  -\textit{MA-SAM} with automatic segmentation significantly outperformed various SOTA 3D medical image segmentation methods and demonstrated efficiency in generalization across datasets. \newline -The prompt mode offers a substantial advantage, particularly for challenging tumor segmentation tasks. \\\hline

2024 \cite{shen2024fastsam3d} & 3D SAM (TS)&-Points. & Three MIS datasets.&  \textit{FastSAM3D}, a lightweight model, offers improved memory efficiency and faster inference compared to 2D and 3D SAMs while maintaining stable performance close to the best methods. \\\hline \hline


2023 \cite{li2023segment}& Semi-Supervised (TS)& -Pseudo Label prompt.& ACDC \cite{bernard2018deep}& Their proposed SSL framework outperformed two baseline approaches when using 5\% of the training data. \\\hline

2024 \cite{zhang2023semisam}   & Semi-Supervised (TS)& -Points from PL.& Left Atrium Segmentation Challenge Dataset \cite{xiong2021global}& -\textit{SemiSAM} outperformed other semi-supervised methods and SAM-based scenarios, particularly in cases with extremely limited labeled samples (1–4).  \\\hline \hline

2024 \cite{yue2024surgicalsam}& Auto Prompting (TS)& Automatic Prompts learning &EndoVis2018 \cite{allan20202018} and EndoVis2017  \cite{allan20192017} &
Comparisons showed that \textit{SurgicalSAM} achieved better than other SOTA segmentation approaches and SAM-based models  in both performance and efficiency.  \\\hline

2025 \cite{lei2025medlsam} &  Auto Prompting (TS)& Auto Bounding box localization   &  WORD \cite{luo2022word} and StructSeg19 Task1 \cite{structseg2019}) encompassing 38 organs. & Experiments  showed that \textit{MedLSAM} achieves comparable performance to promptable SAM-based approaches with manual prompts, significantly reducing reliance on expert input. \\\hline

\end{tabular}}
\end{table*}

\subsection{Zero-Shot SAM for MIS}
The direct approach is to investigate the performance of original SAM trained on natural imaging without any re-training in MIS (zero-shot), to this end plenty of works have investigated its performance for various medical imaging segmentation tasks and modalities which includes and not limited to CT-scan \cite{wald2023sam}, MRI \cite{mohapatra2023sam}, histopathological \cite{deng2023segment}, colonoscopic \cite{zhou2023can}, endoscopic \cite{wang2023sam} and multi-modalities \cite{huang2024segment, mazurowski2023segment}.

In \cite{mazurowski2023segment}, the performance of SAM was evaluated on 19 medical imaging datasets, revealing significant variability across tasks. SAM performed better on well-defined objects (e.g., hip X-ray) and worse on ambiguous cases (e.g., spine MRI), with box prompts yielding notably better results than point prompts. Compared to prompt-click methods like RITM \cite{sofiiuk2022reviving}, SimpleClick \cite{liu2023simpleclick}, and FocalClick \cite{chen2022focalclick}, SAM achieved superior results in most of the evaluated tasks. However, in iterative segmentation scenarios with five or more user-provided points, its superiority diminished, with other methods surpassing its performance. While SAM demonstrates impressive zero-shot capabilities and strong performance in single-point prompt settings, its moderate to poor results in certain scenarios underscore the need for careful application in medical imaging tasks.

In a similar experimental study investigating the zero-shot performance of SAM for medical image segmentation, Y. Huang \textit{et al.} evaluated SAM's performance across extensive datasets constructed from 52 open-source datasets spanning 16 imaging modalities \cite{huang2024segment}. The study investigated three types of prompting: point, box, and automatic everything. Among these, manual prompting demonstrated superior performance compared to the automatic everything mode. Furthermore, SAM showed significant potential for segmenting certain objects and imaging modalities while underperforming for others. These experiments highlight that zero-shot SAM is insufficient to guarantee reliable performance in direct applications. However, it can serve as a valuable starting point, significantly reducing the time required for manual label creation.

\subsection{SAM Fine-Tuning for MIS}
Different from zero-shot evaluation scenario, adopting SAM through transfer learning have been widely investigated, where it could be classified into two main approaches: (i) fine-tuning \cite{li2024polyp, ma2024segment}, and (ii) Parameter-efficient Fine-tuning  (PEFT) \cite{wu2023medical, feng2023cheap, cheng2023Sammed2d, zhang2023customized, paranjape2024adaptivesam, paranjape2024s}.

In \cite{ma2024segment}, J. Ma \textit{et al.} introduced MedSAM, a foundation model for medical image segmentation built on the SAM framework \cite{kirillov2023segment}. It features a ViT-base image encoder trained with the MAE approach \cite{he2022masked}, a box prompt embedding encoder, and a lightweight decoder integrating transformer-based prompt-image fusion and deconvolutional blocks \cite{kirillov2023segment}. MedSAM was pre-trained on the SA-1B dataset for natural image segmentation and fine-tuned on over 1.5 million medical image-mask pairs in an end-to-end manner. MedSAM excels in generalizing to unseen classes and modalities, outperforming conventional segmentation architectures. However, it faces challenges with underrepresented modalities due to imbalanced training data and with segmenting structures like vessels that have weak boundaries and low contrast, where box prompts are less effective. Notably, MedSAM offers a significant advantage in expert annotation, reducing labeling time by up to 82\%.

J. N. Paranjape \textit{et al.} introduced S-SAM, an adaptation of SAM for medical image segmentation that requires training only 0.4\% of SAM's parameters \cite{paranjape2024s}. This approach leverages SVD-based tuning to update the weights of the MSA in SAM's image encoder, significantly reducing the number of trainable parameters. In this setup, only the singular values, positional embeddings, and layer normalization parameters are trainable within the image encoder. Additionally, the authors proposed a text prompt encoder to eliminate the need for expert intervention. This encoder utilizes CLIP as the text embedder, followed by a Text Affine Layer (TAL) and SAM's prompt encoder, with only the TAL parameters being trained. For segmentation, SAM's mask decoder is employed without modifications. S-SAM demonstrated superior performance across five MIS datasets compared to SOTA methods and other SAM-based models, achieving remarkable results with minimal trainable parameters.

More PEFT methods have been explored to fine-tune specific SAM parameters for MIS. In \cite{zhang2023customized}, LoRA \cite{hu2021lora} was applied to the query and value projection layers of SAM's image encoder, with only LoRA parameters being updated, while all other parameters were frozen. Additionally, the mask decoder and prompt encoder were trained, with the prompts are only used during training and not required for inference. In \cite{cheng2023Sammed2d}, J. Cheng \textit{et al.} introduced an adaptation block after each transformer's MHSA block to capture specialized medical imaging features. The adaptation block incorporates channel attention followed by spatial feature enhancement through convolutional down-sampling and up-sampling operations. During training, only the image encoder's adaptation blocks, prompt embedding, and mask decoder are updated.


\subsection{Refined SAM for MIS}
Different from the attempts that have tried to adopt SAM for MIS, other attempts have investigated how to modify SAM's architecture to achieve efficient performance in MIS \cite{zhang2023input, zhang2023sam}.

In \cite{zhang2023input}, Y. Zhang \textit{et al.} demonstrated that SAM model could be leveraged to augment input images and enhance model performance. To achieve this, they proposed SAMAug, a method that utilizes segmentation masks obtained through grid prompts (automatic everything) to generate prior segmentation and boundary maps. These prior maps are concatenated with the grayscale input image to form three-channel inputs, mimicking the structure of an RGB image. The augmented images are then used to train a baseline architecture, such as U-Net \cite{ronneberger2015u}. Experimental results showed that employing SAM for input augmentation significantly improved performance in MIS across three evaluation datasets.

In \cite{zhang2023sam}, J. Zhang \textit{et al.} proposed SAM-Path, a dual-path architecture for histopathology tissue segmentation. This architecture employs two parallel encoders: a pathology encoder \cite{chen2022scaling} and the vanilla SAM image encoder. The features extracted by these encoders are concatenated and passed through a dimensionality reduction module. Instead of relying on manual prompts, SAM-Path adopts the approach introduced in \cite{zhang2023prompt}, using trainable tokens for each class as prompts. These tokens serve as input for the decoder, enabling it to segment each class independently. Experiments conducted on two histopathology datasets demonstrated the effectiveness of combining the domain-specific encoder with the SAM encoder, outperforming various foundational approaches, including fine-tuning SAM.

\subsection{From 2D to 3D SAM for MIS}

Since SAM have been initially proposed for 2D natural imaging segmentation, several works have been proposed for 3D MIS, which can be classified into two main categories: (i) adopting pretrained 2D SAM to 3D \cite{wu2023medical, gong20233dsam, gong20233dsam, bui2024sam3d} and (ii) 3D SAM trained from scratch \cite{wang2024sammed3d}. 

In \cite{wu2023medical}, J. Wu \textit{et al.} introduced Med-SA, an adaptive foundation model for 3D medical image segmentation, with two key innovations: a parameter-efficient fine-tuning strategy and SD-Trans for 3D adaptation. Med-SA integrates LoRA \cite{hu2021lora} blocks in the encoder transformer layers, while the decoder employs both LoRA and Hyper-Prompting Adapters (HyPAdpt). Inspired by hypernetworks \cite{ha2016hypernetworks}, HyPAdpt integrates prompt embeddings by projecting, reshaping, and multiplying them with adapter embeddings for deep feature-level prompting, with only these adaptation blocks updated during training. For 3D adaptation, SD-Trans processes features through two shared-parameter MSA branches, handling spatial and depth dimensions separately, then fuses them via summation. Experimental results showed that Med-SA outperforms MIS state-of-the-art methods and foundation models (SAM \cite{kirillov2023segment} and MedSAM \cite{ma2024segment}), achieving the best performance  with box prompts (0.75 overlap) on most of the evaluated datasets.

Similarly, MA-SAM \cite{chen2024ma} extends the 2D SAM model to 3D medical imaging segmentation by integrating 3D adapters into the encoder's Transformer blocks, positioned before and after the MSA. Each adapter consists of a linear down-projection, a 3D convolutional layer for volumetric feature extraction, and a linear up-projection, significantly reducing trainable parameters. To accommodate SAM's 2D operations, features from the 3D adapters are reshaped by merging the batch and depth dimensions. To minimize trainable parameters of the image encoder, MA-SAM incorporates FacT layers \cite{jie2023fact}, which share most parameters across Transformer layers, eliminating the need to train the MSA parameters of the encoder. The decoder retains SAM's structure but replaces the two (x4) upsampling layers with four (x2) deconvolutional layers and keeping all parameters trainable. Experiments across various MIS tasks and modalities demonstrate MA-SAM's robustness compared to SOTA segmentation architectures without using any prompts. For pancreas tumor segmentation in CT images, which exhibits substantial challenges, MA-SAM achieved outstanding performance using a tight 3D prompting box.

Different from directly adopting pretrained 2D SAM models to 3D, H. Wang \textit{et al.} proposed a foundation model for 3D medical imaging segmentation called SAM-Med3D, trained from scratch \cite{wang2024sammed3d}. SAM-Med3D was trained on large 3D medical imaging datasets called SA-Med3D-140K, which were constructed from 70 public and 24 private datasets consisting of 22K 3D images and 143K corresponding 3D masks across 28 MI modalities. It adopts SAM's components (image and prompt encoders and mask decoder) into 3D versions using 3D convolutions, learnable 3D absolute positional encoding, and 3D attention blocks. SAM-Med3D, with minimal prompts, demonstrated superior performance over 2D SAM adaptations on both seen and unseen targets. Furthermore, SAM-Med3D achieved competitive performance compared with task-specific models using only 1 prompt point per volume and outperformed them in most evaluation datasets with 10 prompt points per volume, excelling on unseen data sources and modalities.

Other works have been investigating different approaches to adopt SAM for 3D MIS.  In \cite{shen2024fastsam3d}, to cope with the high memory and time demands of 3D tasks, FastSAM3D was proposed as a lightweight, accelerated model without significant performance decline. Key contributions include exploring knowledge distillation from a 12-layer ViT-B encoder of SAM-Med3D\cite{wang2024sammed3d} to a 6-layer ViT-Tiny, removing MHS in the first two layers while retaining only the FFN, and replacing MHS with a 3D sparse flash attention scheme in both the encoder and decoder.

\subsection{SAM in Semi-Supervised MIS}
In addition to use SAM for providing the segmentation masks from medical imaging, it has been explored to overcome with annotation cost
limitation for medical imaging to provide pseudo and weakly labels in semi-supervised learning mechanism \cite{li2023segment, zhang2023semisam, zhang2023samdsk}.

In \cite{li2023segment}, N. Li \textit{et al.} integrated SAM into a semi-supervised framework, using a generated pseudo-label as a prompt for SAM to produce more reliable pseudo-labels, thereby enhancing SSL performance. Results on the ACDC dataset \cite{bernard2018deep} demonstrated the effectiveness of this approach, outperforming other methods on 5\% of the training data.
Another approach leveraging SAM's foundational capabilities in a SSL framework is SemiSAM \cite{zhang2023semisam}. It consists of two branches: a student-teacher model based on V-Net and a SAM branch. The student model, a V-Net architecture \cite{milletari2016v}, is trained on labeled data using a supervised loss, while the teacher model is an exponential moving average (EMA) ensemble of the student model. Unlabeled data is processed through both branches, with unsupervised consistency applied between their outputs. To enhance consistency regularization, the student model's predictions are used to generate prompt points via an uncertainty-aware strategy, which are then fed into the SAM branch (SAM-Med3D \cite{wang2024sammed3d}). The outputs from both the student and SAM branches serve as additional supervision signals.
Experiments on the Left Atrium Segmentation Challenge Dataset \cite{xiong2021global} demonstrated that SemiSAM outperformed other semi-supervised methods and SAM-based scenarios, particularly in cases with extremely limited labeled samples (1–4).


\subsection{Towards Auto-Prompt SAM for MIS}
Efficient prompt have shown to be crucial for achieving efficient performance, these prompts include points, boxes and masks. In fact, during testing of most of SAM methods in MIS the prompts were obtained from the testing GT. However, this scenario does not correspond to the real scenario, where expert specialists are needed to provide these prompts. To cope with this limitation and move towards automatic prompts for real application scenarios, much efforts have been spent for this purpose \cite{Pandey_2023_ICCV, lei2025medlsam} \cite{shaharabany2023autosam, yue2024surgicalsam} \cite{deng2023sam}.

In \cite{yue2024surgicalsam}, W. Yue \textit{et al.} proposed SurgicalSAM, an automatic prompt generation method using class prototyping for surgical instrument segmentation. SurgicalSAM consists of an image encoder, a class prototype prompt encoder, and a mask decoder, following the SAM structure. Class-specific regions are identified by computing the similarity between a class prototype bank and the input image's feature embeddings, which are then used to assign prompts. The mask decoder leverages these prompts and image embeddings to segment the target class. During training, only the prompt encoder and mask decoder were fine-tuned using a segmentation loss for mask prediction and a contrastive loss for prototype learning. SurgicalSAM demonstrated superior performance and efficiency compared to state-of-the-art methods and SAM-based models on two surgical instrument segmentation datasets.

To adopt SAM in an automated framework, W. Lei \textit{et al.} proposed MedLSAM, comprising two main components: MedLAM for automated target localization and SAM for segmentation \cite{lei2025medlsam}. MedLAM employs two self-supervised learning tasks to leverage a small annotated support set to identify target anatomical regions in 3D medical images. 2D localizations from each slice are used as bounding box prompts for SAM to segment the target class. Experiments on two datasets (WORD \cite{luo2022word} and StructSeg19 Task1 \cite{structseg2019}) covering 38 organs showed that MedLSAM achieves comparable performance to promptable SAM-based approaches with manual prompts, significantly reducing reliance on expert input.


\subsection{Discussion}

SAM has had a significant impact on the field of medical image segmentation, driving substantial progress. SAM-based approaches have been proposed as foundation models to predict segmentations for both seen and unseen modalities and targets \cite{wald2023sam, mazurowski2023segment, huang2024segment, ma2024segment, cheng2023Sammed2d, wang2024sammed3d}, as well as task-specific models that compete with SOTA methods \cite{zhang2023customized, paranjape2024s, paranjape2024adaptivesam, zhang2023sam, zhang2023input, gong20233dsam, wu2023medical, chen2024ma}.

In zero-shot scenarios, SAM has demonstrated good performance in certain medical imaging modalities and tasks, particularly when objects are well-distinguishable \cite{wald2023sam, mazurowski2023segment, huang2024segment}. However, challenges such as low contrast, ambiguous tissue boundaries, and tiny lesion regions persist. Additionally, imaging modalities that deviate significantly from natural RGB images can impact performance, as the original SAM model lacks medical imaging feature awareness. Among the different prompt types, bounding box prompts have been found to yield better segmentation performance. Fine-tuning SAM as a foundation model has also shown potential but remains limited by ambiguous boundaries and underrepresented imaging modalities during training \cite{ma2024segment}.

Using SAM as a task-specific model within a supervised learning paradigm, where it is trained and evaluated exclusively on seen modalities and targets, has proven effective, outperforming various SOTA methods \cite{zhang2023customized, wu2023medical}. However, these methods often require expert intervention during testing to provide prompts. To address this limitation, several strategies have emerged, including text prompts \cite{paranjape2024s, paranjape2024adaptivesam, shen2024fastsam3d}, auto-prompt learning \cite{zhang2023input}, and even prompt-free approaches \cite{chen2024ma}.

Recent advancements have also demonstrated SAM's effectiveness in semi-supervised learning frameworks, where pseudo-labels are used as prompts to improve performance \cite{li2023segment, zhang2023semisam}. Moreover, automatic prompting methods have been introduced to eliminate the need for expert intervention, further streamlining the process \cite{yue2024surgicalsam, lei2025medlsam}.

In summary, SAM-based approaches hold great promise in the field of medical image segmentation. However, creating an effective foundational model remains challenging, requiring extensive labeled data, computational resources, and coverage of diverse medical imaging modalities. Task-specific SAM adaptations have achieved SOTA performance across numerous tasks, and advancements in automatic prompting are poised to drive further progress. Additionally, SAM foundation models can serve as efficient tools in the labeling process when used in collaboration with experts.


\section{Universal Models}
\label{Sec:UM}
In-context learning, first introduced in models such as GPT-3 \cite{achiam2023gpt}, enables models to adapt to new tasks by leveraging a few examples directly provided as input. This paradigm has been extended to vision models \cite{wang2023seggpt, xu2024eviprompt}, demonstrating its versatility across various tasks. In medical imaging, in-context learning holds promise for advancing the field from task-specific models toward the development of universal models capable of generalizing to diverse tasks. These universal models can effectively handle more challenges with minimal examples \cite{butoi2023universeg, rakic2024tyche}, eliminating the need for retraining or fine-tuning, thus paving the way for more adaptable and efficient solutions in medical image segmentation. Recently, interest in universal models for MIS has grown, as highlighted in \cite{liu2023clip, butoi2023universeg, rakic2024tyche, hu2024icl, xu2024eviprompt}.


\subsection{Problem Formulation}
In the traditional approach to medical imaging segmentation, for a task $k_j$ containing $C_j$ classes, a model $f$ is trained using $N$ samples of \texttt{<image-mask>} pairs $(I_i, Y_i)_{i=1}^N$, where $Y_i$ contains labels for all classes present in $I_i$. The model predicts segmentation masks as $\hat{y_i} = f(I_i)$. For a new class $c_j'$ or task $k_{j'}$, a new model $f'$ is typically trained, or the existing model $f$ is fine-tuned using new data that includes the updated class or task.

\begin{figure}[htp]
\centering
\includegraphics[width = 6.7in, height=2.7in]{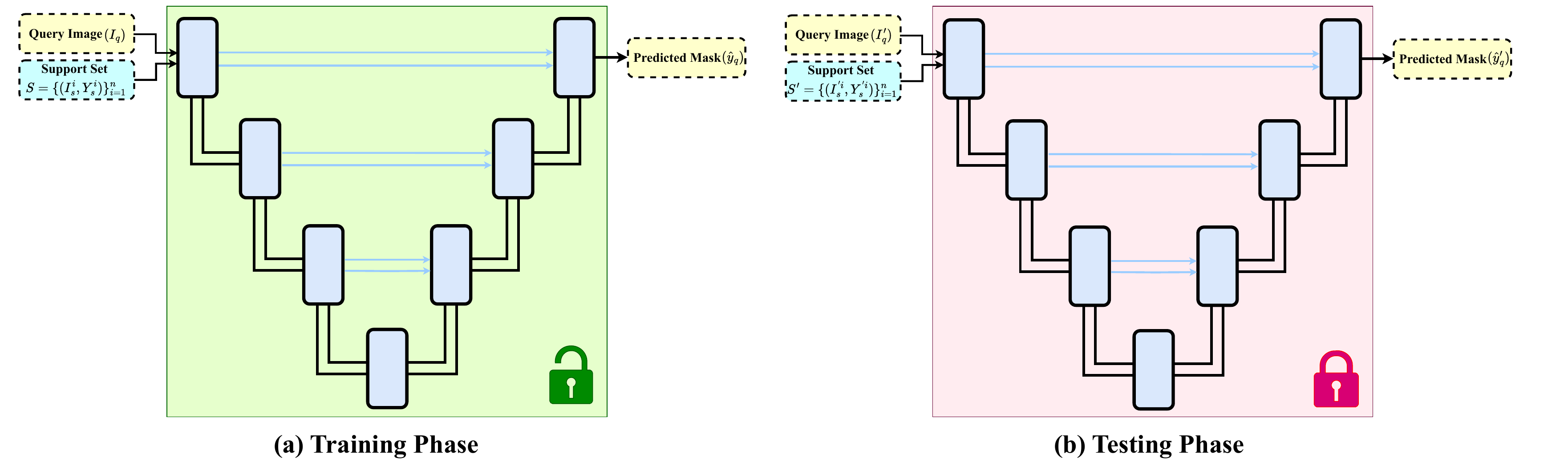}
\caption{General structure of the universal model. During training, a query image \( I_q \) and a support set \( S = \{(I_s^i, Y_s^i)\}_{i=1}^n \) corresponding to task \( k \) are used to train the model \( U \). During inference, a target query image \( I'_q \) and a support set \( S' = \{(I_s^i, Y_s^i)\}_{i=1}^n \) corresponding to a new task \( k' \) (unseen during training) are fed to the model \( U \). The model predicts the segmentation mask \( {y'} \) for \( I'_q \), without any retraining of fine-tuning.}
\label{fig:univ}
\end{figure}

The aim of a universal model is to develop a single model that can generalize across any new class or task, even from the same or different imaging modalities, using only a few samples and without requiring additional training or fine-tuning as depicted in Figure \ref{fig:univ}. In general, a universal model $U$ is capable of predicting the target class $c'$ from a query image $I'_q$ with using a support set $S'$, which contains labeled examples of $c'$. The support set is defined as $S' = \{(I_s^{'i}, Y_s^{'i})\}_{i=1}^n$, where $n$ is the number of examples in the support set. The prediction is then formulated as: $\hat{y}'_q = \mathbb{U}(I'_q, S')$.

\begin{figure}[htp]
\centering
\includegraphics[width = 6.7in, height=2.8in]{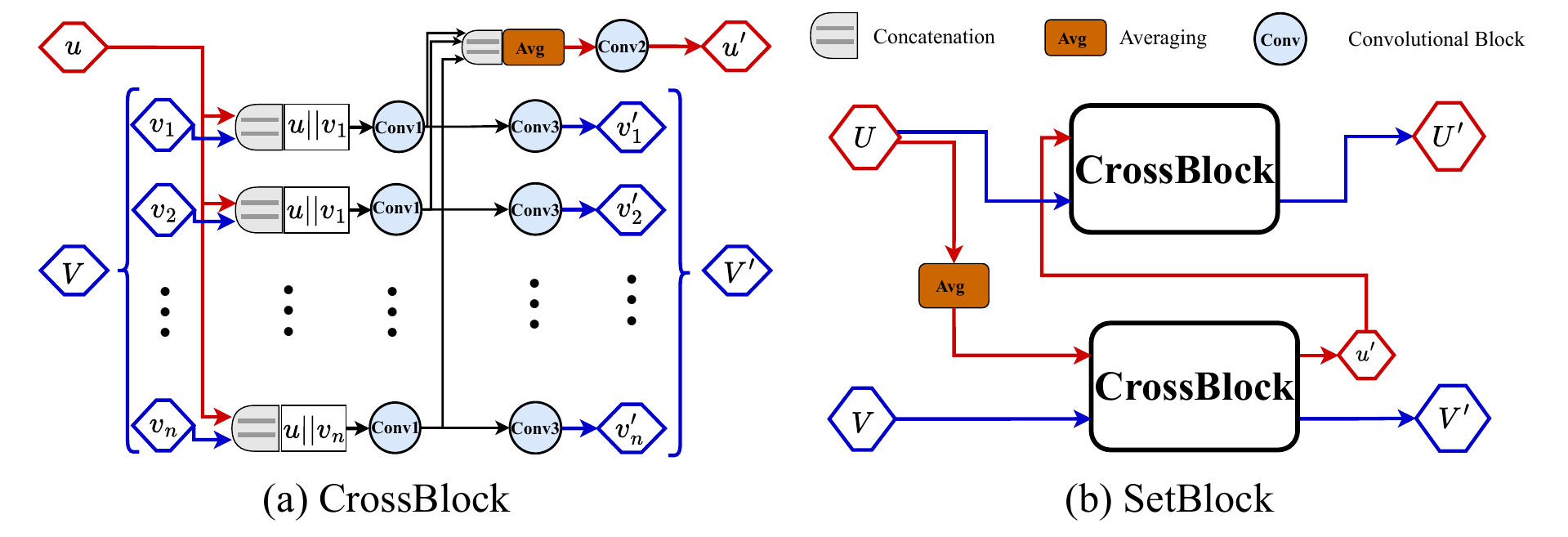}
\caption{Detailed illustration of CrossBlock and SetBlock represented in \cite{butoi2023universeg} and \cite{rakic2024tyche}, respectively. Better view in color, red and blue colors represent query and support paths, respectively.}
\label{fig:univb}
\end{figure}


\subsection{State-of-the-Art of Universal Models in MIS}
In \cite{liu2023clip}, J. Liu \textit{et al.} proposed a Universal model to segment 25 abdominal organs and 6 abdominal tumor types from CT scans in a partial annotation scenario. Their proposed approach, called CLIP-Driven Universal, consists of two paths: class and image embeddings, which are CLIP and an Encoder-Decoder architecture (U-Net), respectively. Both the class embedding and general image representation (the output of the image encoder features) are concatenated and passed through an MLP layer serving as a text-based controller \cite{tian2020conditional} to produce the controlling parameters. These parameters are applied to the segmentation head, which processes the U-Net decoder output features using sequential convolutional blocks. Through masked backpropagation, the weights of the model are updated. Extensive experiments showed the effectiveness of CLIP-Driven Universal for both generalizability and transferability.

In \cite{butoi2023universeg}, V. I. Butoi \textit{et al.} introduced the MegaMedical dataset, derived from 53 MIS datasets spanning diverse anatomies and imaging modalities. They proposed UnivSeg, a U-Net-based architecture designed to segment new classes and generalize across both seen and unseen imaging modalities. UnivSeg comprises their proposed CrossBlock modules (Figure \ref{fig:univb}.a), along with downsampling in the encoder layers and upsampling in the decoder layers. UnivSeg processes a query image $q$ and a support set of concatenated image-mask pairs $S$ as input. As shown in Figure \ref{fig:univb}.a, CrossBlock extracts interaction features between the query $u$ and the support set $V$. First, the query feature maps are concatenated with each of the support feature maps $(v_1, v_2, ..., v_n)$ and then passed through a shared convolutional block $conv_1$. The resulting interaction features are used to obtain the updated query maps and the set of updated support maps by concatenating them and passing them through $conv_2$ for the former, and by iterative refinement through $conv_3$ for each of the support set. Notably, in the first CrossBlock, the inputs $u$ and $V$ correspond to the query image $I_q$ and the set of concatenated support image-mask pairs $S$, respectively. The refined query $u'$ and support maps $V'$ are further processed through subsequent CrossBlocks in UnivSeg until the final prediction mask is obtained. Experiments demonstrated UnivSeg's superiority over FSS approaches and its competitive performance compared to the supervised upper bound (nnU-Net \cite{isensee2021nnu}) across held-out test datasets. Results highlighted the importance of training with diverse datasets and showed that increasing the support set size improves performance. This universal approach is well-suited for segmenting new tasks without retraining or fine-tuning, particularly in low- and medium-annotation regimes.

Building on UniverSeg \cite{butoi2023universeg}, M. Rakic \textit{et al.} introduced Tyche, a universal model for MIS that addresses unseen tasks and prediction uncertainty \cite{rakic2024tyche}. Tyche generates diverse predictions for a target image by using candidate representations, copies of the input image concatenated with a channel of Gaussian noise. Building on CrossBlock \cite{butoi2023universeg}, SetBlock (Figure \ref{fig:univb}.b) is proposed to extract interaction features between the set of candidate representations $U$ and the support set $V$. It comprises two CrossBlocks: the first processes mean candidate representations with the support set to generate updated query $u'$ and support maps $V'$. The second CrossBlock interacts the updated query maps $u'$ of the previous CrossBlock with the candidate representations $U$ to produce the updated candidate maps $U'$.  Thus, the obtained updated query candidates $U'$ and support  $V'$ are consecutively processed through the SetBlocks in a U-Net-like architecture until  segmentation predictions of the input candidates are obtained. The loss function optimizes only the best candidate prediction. Experiments on unseen tasks demonstrated Tyche's superiority over in-context, interaction, and task-specific segmentation methods. Additionally, Toyche was introduced to provide multiple predictions, enabling practitioners to select the most accurate one.

Universal models have demonstrated promising performance in handling domain shifts and adapting to new segmentation tasks with minimal examples \cite{butoi2023universeg, rakic2024tyche}. However, the definition of a universal model can vary significantly across works and often overlaps conceptually with FSS and foundation models. One notable limitation, as this field is still in its infancy, is the lack of clear benchmark datasets for comparing the performance of universal models. This is largely due to privacy constraints that prevent the re-sharing of medical datasets, which can hinder progress in advancing universal models for MIS. Recent advancements propose combining universal models with foundation models to enhance adaptability and performance in complex scenarios \cite{liu2023clip, hu2024icl, xu2024eviprompt}. Overall, the concept of a universal model remains flexible, evolving to meet the requirements of specific tasks and applications.

\section{Discussion}
\label{Sec:dis}
The field of medical image segmentation has witnessed significant advancements with the adoption of SOTA vision approaches such as generative models, few-shot learning, foundation models, and universal models. Each of these paradigms has introduced innovative methods to address the inherent challenges in MIS, including limited data availability, domain shifts, task complexity, and model scalability. However, there remain several critical issues that require attention to fully realize the potential of these methods in real-world applications.

Generative models have revolutionized medical image segmentation, advancing beyond traditional data augmentation into areas like semi-supervised and self-supervised learning. Approaches based on GANs, diffusion models, and hybrid techniques have been applied in various scenarios, including data augmentation, cross-modality learning, and segmentation. However, the absence of a unified evaluation protocol and dedicated benchmark datasets hinders the effective comparison of these models. Establishing standardized evaluation frameworks and encouraging the release of reproducible implementations would significantly benefit this domain. Furthermore, integrating generative models with foundation or universal models, such as large vision transformers or pre-trained self-supervised networks, could enhance performance, addressing the limitations of individual methodologies.

Few-shot segmentation has shown remarkable progress, particularly in scenarios where labeled data is scarce. Prototype networks and hybrid methods leveraging contextual relationships, such as 3D models, have demonstrated encouraging results. However, the field still suffers from limited task diversity and inconsistent evaluation protocols. Thus, future works need to explore diversity tasks and classes, as well as more consistent evaluation protocols. Furthermore, efforts to narrow the performance gap between FSS and fully supervised learning through innovative architectures or training
paradigms will be crucial for the widespread adoption of FSS in medical practice.

Foundation models like SAM have demonstrated remarkable versatility in segmenting both seen and unseen targets across different modalities, paving the way for broader applications in MIS. While their zero-shot performance has been impressive in some scenarios, it remains limited in others. To address these challenges, various approaches have been explored, including fine-tuning SAM, employing task-specific adaptations, developing 3D SAM models, and integrating semi-supervised learning frameworks. These methods have shown promise in advancing foundation models for MIS. However, significant challenges persist, such as low contrast, ambiguous boundaries, multiple disconnected regions of interest, and underrepresented imaging modalities. Enhancing the generalizability of these models remains crucial, particularly for modalities and tasks that diverge significantly from natural image characteristics. Currently, most existing foundation models have not incorporated the refinement-based interaction strategies used during SAM’s original training. Integrating such interactive mechanisms could greatly improve SAM’s adaptability for MIS. Furthermore, deploying SAM-based platforms that allow direct interaction with medical professionals, coupled with active learning or reinforcement learning, could maximize SAM’s potential in clinical settings. While SAM has proven to be a strong starting point for generating initial segmentation masks and accelerating the annotation process, developing lightweight and user-friendly adaptations is a fundamental step toward integrating SAM into MIS and enabling broader use by medical practitioners.

Universal models offer a promising avenue for handling domain shifts and adapting to new tasks with minimal examples. Despite their potential, the lack of clear definitions and standardized benchmarks poses a significant challenge, particularly in the context of medical imaging, where privacy constraints limit dataset availability. Collaborative efforts to establish open, privacy-compliant benchmarks could accelerate the progress and enable meaningful comparisons across universal model implementations. Moreover, privacy constraints limiting data sharing present significant barriers to benchmarking these
models. Future efforts should explore synthetic datasets or federated learning frameworks to enable rigorous evaluation while ensuring compliance with data privacy regulations.

Overall, while significant strides have been made, the lack of standardized benchmarks, task diversity, and reproducible implementations remains a persistent bottleneck across all paradigms. Collaborative efforts among researchers, clinicians, and policymakers are crucial to overcoming these barriers. By addressing these challenges, the field can pave the way for more robust, scalable, and generalizable MIS solutions, ultimately enhancing their real-world applicability in medical diagnostics and treatment planning. This survey highlights the need for collaborative and
multidisciplinary efforts to overcome the current limitations in MIS. By building on the methodologies and insights discussed here, future research can drive significant advancements in medical image segmentation, ultimately improving diagnostic accuracy and clinical outcomes.


\section{Conclusion}
\label{Sec:Conc}
This survey highlights the significant advancements in MIS through the use of generative models, few-shot learning, foundation models, and universal models. While these approaches show great potential in addressing key challenges such as limited data, domain shifts, and scalability, there remain obstacles such as the lack of standardized benchmarks, privacy constraints, and performance gaps in complex tasks. Future research should focus on refining these models, enhancing their adaptability across diverse modalities, and establishing unified evaluation protocols to further advance MIS and improve clinical applications. The future of MIS looks promising, with continued innovation and collaboration driving progress in the field.



\end{document}